\documentclass[conference]{IEEEtran}
\IEEEoverridecommandlockouts
\usepackage{cite}
\usepackage{amsmath,amssymb,amsfonts}
\usepackage{algorithmic}
\usepackage{graphicx}
\usepackage{textcomp}
\usepackage{xcolor}
\usepackage{orcidlink}
\usepackage{url}
\usepackage{hyperref}
\usepackage{subcaption}
\usepackage{textcomp}
\usepackage{titlesec}
\usepackage{acronym}
\usepackage{changepage}
\usepackage{multirow}
\usepackage{booktabs}
\usepackage{tabularx}
\usepackage{adjustbox}
\usepackage{wrapfig}
\usepackage{comment}
\usepackage[utf8]{inputenc}
\titlespacing{\subsection}{0pt}{12pt plus 2pt minus 2pt}{0pt}
\def\BibTeX{{\rm B\kern-.05em{\sc i\kern-.025em b}\kern-.08em
    T\kern-.1667em\lower.7ex\hbox{E}\kern-.125emX}}
\begin{document}

\title{The State of Lithium-Ion Battery Health Prognostics in the CPS Era}



\author{\IEEEauthorblockN{Gaurav Shinde\IEEEauthorrefmark{1},   Rohan Mohapatra \IEEEauthorrefmark{2}, Pooja Krishan \IEEEauthorrefmark{2},
Harish Garg\IEEEauthorrefmark{3}, Srikanth Prabhu\IEEEauthorrefmark{4}, Sanchari Das\IEEEauthorrefmark{5}, \and Mohammad Masum\IEEEauthorrefmark{6} and Saptarshi Sengupta\IEEEauthorrefmark{2}}

\IEEEauthorblockA{\IEEEauthorrefmark{1}Department of Engineering, San Jose State University, San Jose, CA, USA \\
\IEEEauthorblockA{\IEEEauthorrefmark{2}Department of Computer Science, San Jose State University, San Jose, CA, USA \\
\IEEEauthorrefmark{3}School of Mathematics, Thapar Institute of Engineering \& Technology, Patiala 147004, Punjab, India\\
\IEEEauthorrefmark{4}Department of Computer Science and Engineering, Manipal Institute of Technology, Manipal, KA, India\\
\IEEEauthorrefmark{5}Department of Computer Science, University of Denver, Denver, CO, USA\\}
\IEEEauthorrefmark{6}Department of Applied Data Science, San Jose State University, San Jose, CA, USA \\
Email: 
\IEEEauthorrefmark{1}gauravyeshwant.shinde@sjsu.edu,
\IEEEauthorrefmark{2}rohan.mohapatra@sjsu.edu,
\IEEEauthorrefmark{2}pooja.krishan@sjsu.edu,
\IEEEauthorrefmark{3}harish.garg@thapar.edu, \\
\IEEEauthorrefmark{4}srikanth.prabhu@manipal.edu,
\IEEEauthorrefmark{5} sanchari.das@du.edu,
\IEEEauthorrefmark{6}mohammad.masum@sjsu.edu,
\IEEEauthorrefmark{2}saptarshi.sengupta@sjsu.edu}
}


\maketitle
\begin{abstract}
Lithium-ion batteries (Li-ion) have revolutionized energy storage technology, becoming integral to our daily lives by powering a diverse range of devices and applications.Their high energy density, fast power response, recyclability, and mobility advantages have made them the preferred choice for numerous sectors. This paper explores the seamless integration of Prognostics and Health Management within batteries, presenting a multidisciplinary approach that enhances the reliability, safety, and performance of these powerhouses. Remaining useful life (RUL), a critical concept in prognostics, is examined in depth, emphasizing its role in predicting component failure before it occurs. The paper reviews various RUL prediction methods, from traditional models to cutting-edge data-driven techniques. Furthermore, it highlights the paradigm shift toward deep learning architectures within the field of Li-ion battery health prognostics, elucidating the pivotal role of deep learning in addressing battery system complexities.Practical applications of PHM across industries are also explored, offering readers insights into real-world implementations.This paper serves as a comprehensive guide, catering to both researchers and practitioners in the field of Li-ion battery PHM.
\end{abstract}

\vspace{0.5cm}

\section{Introduction}
\subsection{Lithium-ion Batteries}

\ac{Li-ion} batteries have been a groundbreaking advancement in energy storage technology. These rechargeable powerhouses have become ubiquitous in our modern world, powering a wide array of devices that have become integral to our daily lives. With the ability to store electrical energy efficiently and reliably, it makes them the preferred choice for a multitude of applications. They have the advantages of high energy density, fast power response, recyclability, and convenient to movement, which are unsurpassed by other energy storage systems \cite{9351160}. From smartphones and laptops to electric vehicles, spacecraft, and medical devices, \ac{Li-ion} batteries provide the energy needed to keep these technologies running smoothly. We explore how the concept of \ac{PHM} seamlessly integrates with the diverse applications of \ac{Li-ion} batteries.

\subsection{Prognostics and Health Management(PHM)}
\ac{PHM} is a multidisciplinary approach that combines data acquisition, diagnostics, and prognostics to enhance the reliability, safety, and performance of complex systems. It is a systematic method used to monitor the condition of systems, detect anomalies or faults, assess their severity, and predict their \ac{RUL}.\ac{PHM} leverages a combination of sensor data, machine learning algorithms, and domain expertise to provide real-time insights into the health and performance of systems. This proactive approach enables maintenance and operational decisions to be made based on the actual condition of the system, ultimately reducing downtime, minimizing maintenance costs, and improving overall operational efficiency.

\begin{figure}[htbp]
    \centering
    \includegraphics[width=\linewidth]{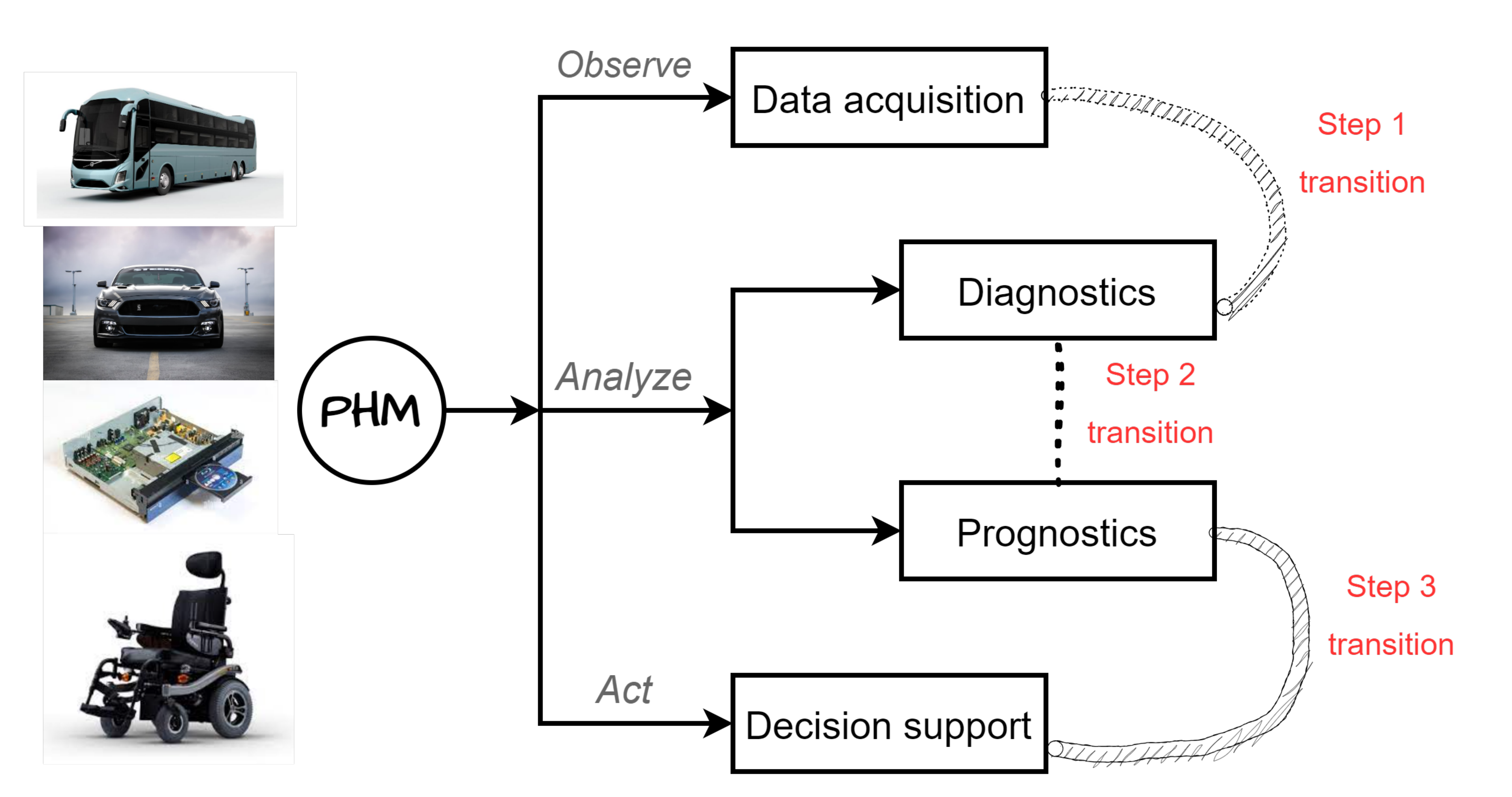}
    \caption{Prognostics and health management framework}
    \label{Prognostics and health management framework}
\end{figure}

\vspace{0.15 cm}
\subsubsection{\textbf{Remaining Useful Life}}
\vspace{0.15 cm}
Remaining Useful Life (\ac{RUL}) is a concise term representing the estimated duration a system or component will remain operational before reaching a critical failure point. It's a crucial concept in prognostics. The \ac{RUL} is currently one of the most important parameters to predict the failure of a component before it effectively occurs \cite{83,84}. It is predicted by detecting the components' conditions through data gathering (e.g., vibration signals) from monitoring sensors \cite{85} and involves adopting model-based (or physics-based), data-driven, or hybrid methodologies. This \ac{RUL} metric plays a key role in guiding maintenance and replacement decisions.
   Below is a generalized equation that models \ac{RUL} of a system.

\vspace{0.15 cm}
\begin{center}
\( RUL = F(C, T, \Delta S) \)
\end{center}

Where:
\begin{itemize}
    \item $RUL$ is the Remaining Useful Life of the system.
    \item $F$ represents a function that models the relationship between the condition of the system and its remaining useful life.
    \item $C$ stands for the current condition of the system, which could be assessed through direct inspections, or sensor data.
    \item $T$ denotes the operational time or age of the system. This could be the total operational hours, number of charge-discharge cycles, or any time-based measure relevant to the degradation of the system.
    \item $\Delta S$ represents the change in system performance or state over time.
\end{itemize}

\vspace{0.15 cm}
\subsection{\textbf{Why is PHM so important?}}
\vspace{0.15 cm}
Prognostics and Health Management (\ac{PHM}) plays a pivotal role in ensuring the reliability and longevity of critical systems, and its importance cannot be overstated. Here are several key reasons why \ac{PHM} is so essential:
\begin{enumerate}
    \item \textbf{Minimizing Downtime:} \ac{PHM} allows for early detection of system anomalies and impending failures. By addressing issues before they escalate, organizations can significantly reduce unplanned downtime, ensuring continuous operation and productivity.
    \item \textbf{Cost Reduction:} Timely maintenance and repair based on \ac{PHM} insights can prevent catastrophic failures and costly emergency repairs. This proactive approach often results in substantial cost savings over reactive maintenance strategies.
    \item \textbf{Enhanced Safety:} In safety critical industries such as aerospace and healthcare, \ac{PHM} ensures the safe operation of systems and equipment. Detecting potential failures before they occur can prevent accidents and protect human lives.
    \item \textbf{Resource Optimization:} \ac{PHM} helps organizations allocate resources efficiently by focusing on systems that require immediate attention.This prevents over-maintenance of healthy systems and directs resources where they are most needed.
    \item \textbf{Sustainability:} By extending the useful life of systems and reducing waste through targeted maintenance, \ac{PHM} contributes to sustainability goals by minimizing the disposal of prematurely retired equipment.
    \item \textbf{Competitive Advantage:} Companies that implement \ac{PHM} gain a competitive edge by ensuring the reliability of their products and services, which is particularly valuable in industries where system downtime or failures can result in financial losses and damage to reputation.
\end{enumerate}

Incorporating \ac{PHM} into the management of complex systems, such as those powered by lithium-ion batteries, is not only a prudent approach but also a strategic one that can lead to improved operational efficiency and long-term success.

\vspace{0.15 cm}
\subsection{\textbf{Review Methodology}}
\vspace{0.15 cm}
In its current form, this paper endeavors to curate a notable collection of contributions from researchers in the field of lithium-ion battery prognostics and health management (\ac{PHM}). It's essential to recognize that this compilation, while comprehensive, cannot fully encapsulate the intricate landscape of \ac{PHM} for lithium-ion batteries. Nevertheless, understanding the complexity of this evolving field, our aim remains to provide readers with a concise overview of significant scholarly works spanning various dimensions, all within the confines of a single document.

\hspace{0.2 cm}\textbf{Contributions:}
The major contributions of this paper are:
\begin{enumerate}
\item \textbf{Comprehensive exploration of \ac{RUL} methods:}
This paper delves deeply into the spectrum of Remaining Useful Life (\ac{RUL}) prediction methods, ranging from traditional model-based approaches to contemporary data-driven techniques. By providing an extensive analysis and explanation of these methods, it equips readers with a thorough understanding of the evolution in \ac{RUL} prediction, aiding researchers and practitioners in choosing the most suitable methodology for their specific applications.

   \item \textbf{Insight into the transition to Deep Learning:}
  One of the focal points of this paper is the elucidation of the paradigm shift toward deep learning architectures in lithium-ion battery health prognostics. By offering a compelling rationale for this transition,it guides readers in comprehending the pivotal role of deep learning in addressing the complexities of battery systems. This insight empowers researchers and engineers to harness the potential of deep learning for enhanced battery management.

\item \textbf{Practical applications across industries:}
  This paper goes beyond theoretical discussions and extends its reach to diverse industrial applications of lithium-ion battery health prognostics.It guides readers by presenting relevant material, helping them understand where to commence their review of practical applications.
\end{enumerate}

\begin{figure*}[!ht]
  \centering
      \includegraphics[width=0.9\textwidth]{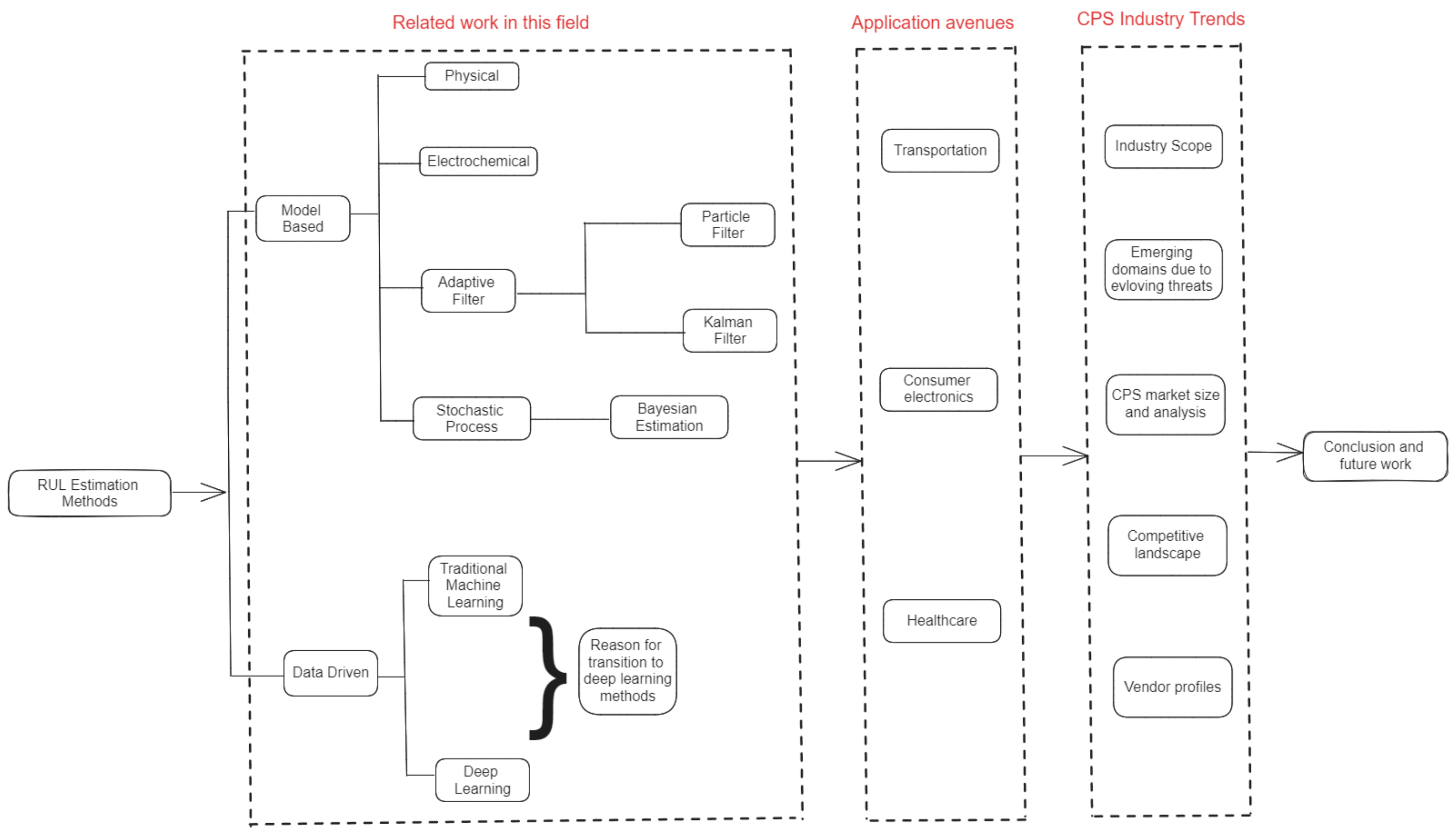}
      \captionsetup{justification=centering}
      \caption{Organization of the paper}
    \label{Organization of the paper}
    \vspace*{-0.1 in}
\end{figure*}

Figure \ref{Organization of the paper} offers a bird's-eye view of the paper's structure.

The paper unfolds as follows: Section II delves into the realm of \ac{RUL} estimation methods, providing a comprehensive overview. Section III discusses industrial applications, whereas sections IV and V explore \ac{CPS} and delve into it's industrial insights respectively. Finally, section VI concludes the paper.
\vspace{0.15 cm}

\section{RUL Estimation Methods}
\vspace{0.15 cm}
In the upcoming section of this research paper, we will conduct a methodical and comprehensive evaluation of various approaches. Each method will be meticulously explained, emphasizing its advantages and limitations. Furthermore, we will provide references to studies that have employed these methods in similar contexts, serving as valuable references for readers seeking further information in this area.

\vspace{0.15 cm}
\subsection{\textbf{Model Based}}
\vspace{0.15 cm}
Model-based methods attempt to set up mathematical or physical models to describe degradation processes, and update model parameters using measured data \cite{11}\cite{12}.Model-based methods could incorporate both expert knowledge and real-time information. Consequently, they may work well in \ac{RUL} prediction of batteries \cite{13}.

However, model-based approaches require accurate knowledge of the battery's internal structure and operating conditions, which can be challenging to obtain in practice. Additionally, these models may not generalize well to different battery chemistries or operating conditions. With this understanding as a foundation, we will now delve into an exploration of different model-based methods.

\vspace{0.15 cm}
\subsubsection{\textbf{Physical Model}}
\vspace{0.15 cm}
Physical model seeks to quantify the factors that influence battery’s performance, and obtains the description of performance evolution. However, considering that several factors may interact with each other to impact the degradation of the performance, it is not easy to make a reliable and precise model to simulate battery system. Thus, this approach usually focuses on the specific physical and chemical phenomena occurring during the utilization's \cite{14}. Physics-based models rely on mathematical equations to describe the battery's physical attributes and controlling principles. These models can give deep insights into the underlying causes of battery deterioration and can be used to predict the behavior of the battery under various operating conditions. However, accurate information about the battery's internal state variables and operating settings is required for these models, which can be difficult to get in practice. Despite these limitations, physics-based models continue to be a significant tool for lithium-ion battery \ac{RUL} estimation, particularly in situations where the operating circumstances are well-defined and the battery's internal state variables can be monitored with high accuracy.

\vspace{0.15 cm}
The physical model of a system uses differential equations to represent the dynamics based on physical laws:
\begin{equation}
    \frac{d\mathbf{x}(t)}{dt} = A(\mathbf{x}(t), \mathbf{u}(t), \mathbf{p})
\end{equation}

Where:
\begin{itemize}
    \item $\mathbf{x}(t)$ is the state vector of the system at time $t$.
    \item $A$ is a vector function representing the system dynamics.
    \item $\mathbf{u}(t)$ is the input or control vector at time $t$.
    \item $\mathbf{p}$ is a vector of the system parameters.
\end{itemize}

\vspace{0.15 cm}\
\subsubsection{\textbf{Electrochemical Model}}
\vspace{0.15 cm}
Electrochemical models are based on precise mathematical models of the electrochemical processes that occur within the battery, such as chemical reactions, lithium-ion and electron movement, and heat impacts. The electrochemical model can represent the internal variables of the battery well, so it can achieve high accuracy \cite{15}. However, due to the complexity and non-linearity of battery behavior, as well as the challenge of precisely describing the electrochemical processes within the battery, establishing accurate electrochemical-based models can be difficult. Furthermore, correct knowledge of the battery's internal state variables \cite{171}, such as the state of charge, temperature, and electrode degradation is required for these models which can be difficult to get in practice. Ongoing research in this field is aimed at producing more accurate and efficient electrochemical models that may be utilized in real-time applications to improve the dependability and safety of lithium-ion batteries.

\vspace{0.15 cm}
For a generalized electrochemical system \cite{172}, the model involves equations that describe charge movement and reaction kinetics:
\begin{equation}
    \frac{\partial c}{\partial t} = D\nabla^2c - \nabla \cdot (\mathbf{v}c) + \mathcal{R}(c, \phi)
\end{equation}
\begin{equation}
    \nabla \cdot (\sigma \nabla \phi) = -Q
\end{equation}

Where:
\begin{itemize}
    \item $c$ is the concentration of the reactive species.
    \item $D$ is the diffusion coefficient.
    \item $\mathbf{v}$ is the velocity field for convective transport.
    \item $\mathcal{R}$ is the reaction rate, which is a function of concentration $c$ and electric potential $\phi$.
    \item $\sigma$ is the conductivity of the medium.
    \item $Q$ represents the source or sink of charge.
\end{itemize}

\vspace{0.15 cm}
\subsubsection{\textbf{Adaptive Filter}}
\vspace{0.15 cm}
The adaptive filter is a digital filter whose coefficient varies with the target in order for the filter to converge to the optimal state \cite{32}. The cost function is the optimization criteria, and the most common is the root mean square of the error signal between the adaptive filter output and the intended signal \cite{33}. Given its ability to detect changes in system dynamics over time, the adaptive filter is a common technique for estimating \ac{RUL} of\ac{Li-ion} batteries. Based on the cost function, the adaptive filter coefficients are changed in real time. Multiple research studies have demonstrated that these adaptive filters function well in \ac{RUL} estimation of \ac{Li-ion} batteries.This section covers the survey of Kalman filter \cite{173} and particle filter \cite{174}, which are popular methods used for \ac{Li-ion} battery \ac{RUL} estimation.

\vspace{0.15 cm}
An adaptive filter updates its coefficients to minimize the error between the predicted output and the actual output. This can be described by the following equations:
\begin{equation}
    \mathbf{y}(t) = \mathbf{H}(t)\mathbf{x}(t)
\end{equation}
\begin{equation}
    \mathbf{e}(t) = \mathbf{d}(t) - \mathbf{y}(t)
\end{equation}
\begin{equation}
    \mathbf{H}(t + 1) = \mathbf{H}(t) + \mu\mathbf{e}(t)\mathbf{x}^T(t)
\end{equation}

Where:
\begin{itemize}
    \item $\mathbf{y}(t)$ is the output of the filter at time $t$.
    \item $\mathbf{H}(t)$ is the matrix of filter coefficients at time $t$.
    \item $\mathbf{x}(t)$ is the input signal vector at time $t$.
    \item $\mathbf{e}(t)$ is the error signal vector at time $t$.
    \item $\mathbf{d}(t)$ is the desired signal vector at time $t$.
    \item $\mu$ is the adaptation rate or step-size parameter.
\end{itemize}

\vspace{0.15 cm}
\paragraph{Kalman Filter}
\vspace{0.15 cm}
The Kalman filter is an adaptive algorithm that iteratively updates a set of equations based on noisy measurements to estimate the state of a dynamic system. It functions in two phases: prediction and update. The prediction phase estimates the system state and covariance matrix based on the previous state and a model of the system dynamics, while the update phase refines the estimate based on fresh measurements. The key benefit of utilizing the Kalman filter for \ac{Li-ion} battery \ac{RUL} estimation is its ability to handle noisy observations and system dynamics, resulting in precise and dependable forecasts. It also gives a recursive estimate of the system state, which makes it appropriate for online estimation applications \cite{173}.The adaptive filter methods have been widely used for \ac{RUL} estimation of \ac{Li-ion} batteries, and several studies have proposed various algorithms based on Kalman filter and its variants. Zheng et al.\cite{34} proposed a combination of \ac{KF} and \ac{RVM} to capture the degradation properties, which showed promising results in \ac{RUL} estimation. Li et al.\cite{35} and Li et al. \cite{36} also proposed \ac{UKF} based methods to predict the \ac{RUL} of \ac{Li-ion} batteries, which achieved good performance on various datasets. Xue et al.\cite{37} proposed an \ac{AUKF} algorithm to adaptively update the process noise covariance and the observed noise covariance, which improved the prediction accuracy and robustness of the filter. Park et al.\cite{38} proposed a hybrid prediction method based on the combination of the KAF -based prediction model and the DGM based \ac{UKF} algorithm, which provided a fast and accurate \ac{RUL} estimation of \ac{Li-ion} batteries. By dealing with noisy and dynamic data and adjusting to changing system circumstances, these approaches give accurate and dependable predictions. In conclusion, the Kalman filter and its variants have proven to be invaluable tools in \ac{Li-ion} battery \ac{RUL} estimation, offering robust and adaptive solutions for handling noisy data and dynamic system conditions.

\vspace{0.15 cm}
\paragraph{Particle Filter}
\vspace{0.15 cm}

\begin{figure}[htbp]
    \centering
    \includegraphics[width=0.9\columnwidth]{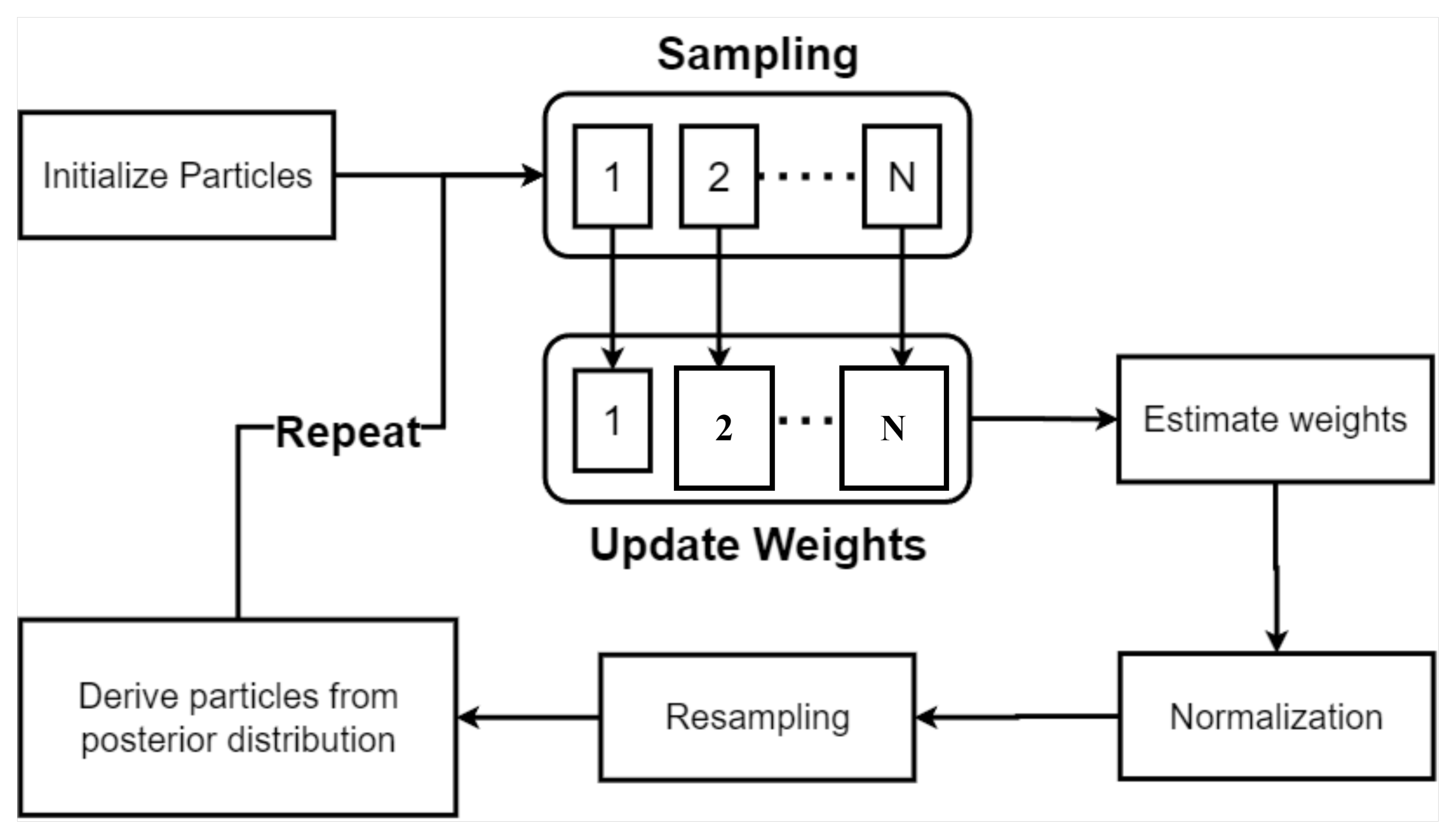}
    \caption{Particle filter process}
    \label{Particle Filter Workflow}
\end{figure}

Particle filtering uses particle sets to represent probability and can be used in any form of the state-space model.The core idea is to express the distribution of random state particles drawn from the posterior probability. In the context of Particle Filtering, the depicted diagram \ref{Particle Filter Workflow} \cite{80} presents a step wise process. It begins with the creation of the Particle Filter and the specification of parameters for the non-linear system, providing a foundation for subsequent operations. The Particle Filter initialization step is crucial, as it sets the initial states for the particles. Following this, the 'Sample Particles' phase generates a representative set of particles for state estimation. The system then acquires measurements, and the subsequent step refines the estimation process.There is a Re-sample decision point, which dictates the course of action:either particles are re-sampled to increase accuracy , or the system proceeds directly to predict the next state. This dynamic approach to particle filtering optimizes state estimation in non-linear systems.Xian et al.\cite{39} proposed a method based on the Verhulst Model, Particle Swarm Optimization and Particle Filter, where particle filter is used to update the predicting model and compensate for the prediction error, thereby improving the accuracy of \ac{RUL} estimation. Hu et al.\cite{40} proposed an integrated method for the capacity estimation and \ac{RUL} prediction of \ac{Li-ion} battery, where the Gauss-Hermite particle filter technique is used for projecting the capacity fade to the \ac{EOS} value. Omariba et al. \cite{41} proposed a Particle Filter method for \ac{RUL} estimation of Li-ion batteries, while Sun et al.\cite{42} used a combination of capacitance, resistance, and constant current charging time with the beta distribution function for \ac{RUL} prediction. Zhang et al.\cite{43} developed a fusion technology composed of a correlation vector machine and a \ac{PF} for \ac{RUL} estimation.In summary, Particle Filter is an effective and widely used technique in the field of \ac{RUL} prediction.

\vspace{0.15 cm}
\subsubsection{\textbf{Stochastic Process}}
\vspace{0.15 cm}
Stochastic process methods are based on statistical theory and combined with other mathematical principles \cite{44}. Stochastic process methods offer a different approach for estimating the \ac{RUL} of \ac{Li-ion} batteries. These are based on the notion that battery degradation is a stochastic process that can be modeled using probabilistic methods. The advantage of stochastic process approaches is that they can represent the unpredictability and uncertainty inherent in the battery deterioration process, but they may need more complicated modeling and computational techniques. With advancements in data-driven and probabilistic modeling, stochastic process approaches may become an increasingly important tool for \ac{RUL} estimation in the future. This section will cover the Bayesian Estimation for remaining useful life prediction of \ac{LIB's}.

\vspace{0.15 cm}
A stochastic process can be described by \ac{SDE} that models the system's evolution over time with a component of randomness. The generalized form of an \ac{SDE} is given by:

\begin{equation}
    dX(t) = f(X(t), t)dt + g(X(t), t)dW(t)
\end{equation}

Where:

\begin{itemize}
    \item \( X(t) \) is the state vector of the system at time \( t \).
    \item \( f(X(t), t) \) is a deterministic function representing the drift of the system, which dictates the direction and speed of the process.
    \item \( g(X(t), t) \) is a function representing the diffusion of the system, which models the uncertainty.
    \item \( dW(t) \) is the incremental process, which introduces the stochasticity into the system.
\end{itemize}

\vspace{0.15 cm}
The specific form of the functions \( f \) and \( g \) will depend on the particular system being modeled. Parameters within these functions can be estimated with techniques such as maximum likelihood estimation or Bayesian inference.

In addition to the SDE, boundary conditions or initial conditions are often specified:

\begin{equation}
    X(t_0) = X_0
\end{equation}

Where \( X_0 \) is the known state of the system at the initial time \( t_0 \). The solution of the SDE, \( X(t) \), provides a probabilistic description of the system's state at any future time \( t \).

\vspace{0.15 cm}
It is also common to include terms for jump processes or to use more complex forms of stochastic calculus to model specific phenomena within the stochastic system.

\vspace{0.15 cm}
\paragraph{Bayesian Estimation}
\vspace{0.15 cm}
Bayesian estimate is a probabilistic method for updating previous knowledge about a system based on new information. It has been used for the \ac{RUL} estimation of \ac{Li-ion} batteries by estimating the probability distribution using Bayesian inference. Bayesian approaches, which may include several sources of information and provide a framework for dealing with uncertainty, are a valuable tool for remaining useful life estimation. This part discusses the different Bayesian estimating approaches that have been proposed for \ac{Li-ion} battery \ac{RUL} estimation. Zhou et al.\cite{45} introduced a unique technique for \ac{RUL} prediction that combines the SBL strategy, which provides multi-step prediction via integrating sub-models rather than direct iterative computing. Refer \cite{45}. Lu et al.\cite{46} suggested a fusion model for uncertainty integration in\ac{RUL} estimation based on \ac{BMA} Refer \cite{46}. Zhao et al. \cite{47} proposed a Bayesian simulator evaluation theory-based update method for the \ac{Li-ion} battery \ac{RUL} prediction model.Finally Zhang et al. \cite{48} proposed a Bayesian Mixture Neural Network for \ac{RUL} estimation that provides reliable prediction intervals.For further information go through \cite{48} .The advancements in Bayesian methods for \ac{RUL} estimation of \ac{LIB's} continue to be an active area of research and further improvements in these techniques are expected to enhance their effectiveness and practicality.

\begin{figure}[htbp]
    \centering
    \includegraphics[width=\linewidth]{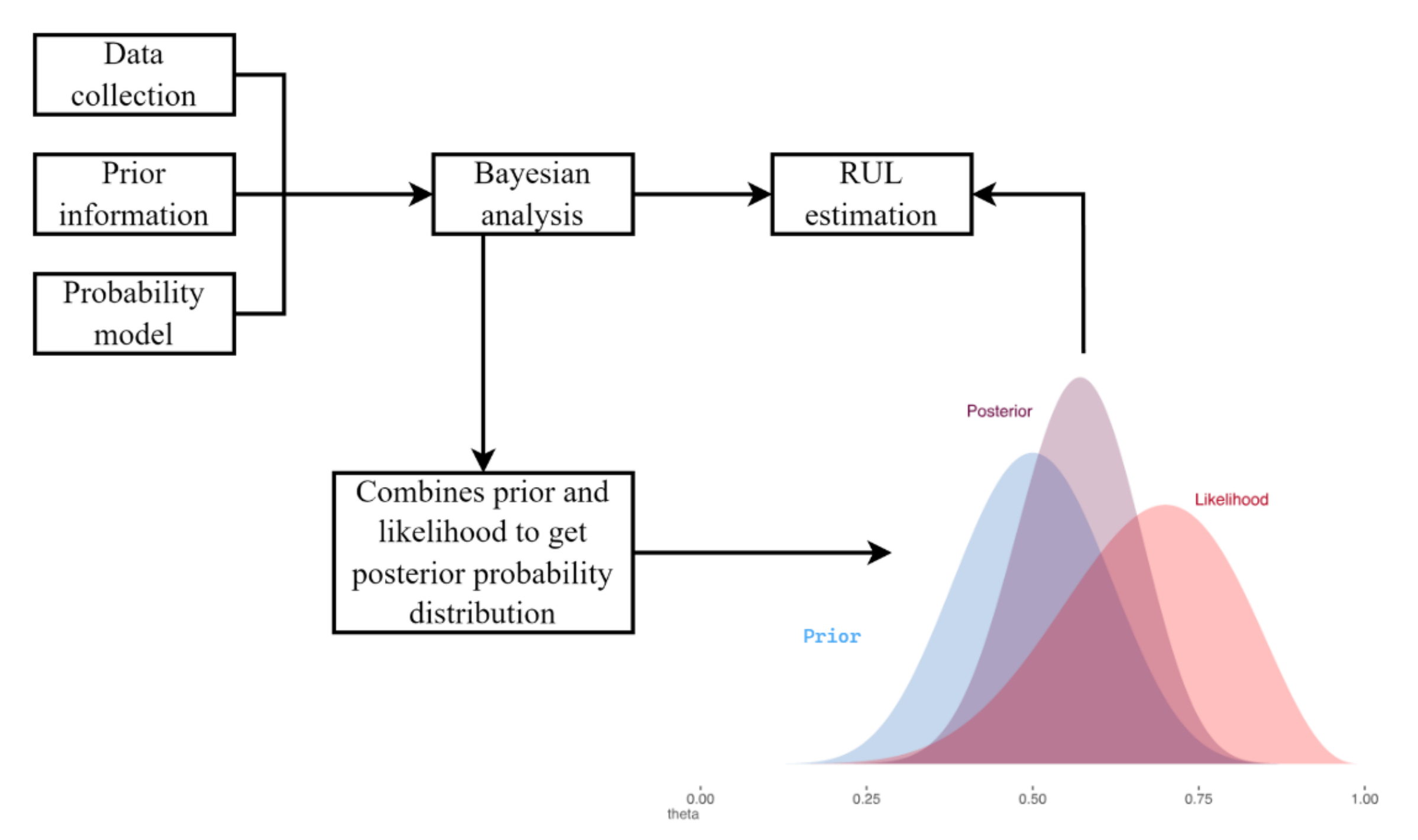}
    \caption{Bayesian estimation process}
    \label{Bayesian estimation process}
\end{figure}

\vspace{0.15 cm}
\subsection{\textbf{Data Driven}}
\vspace{0.15 cm}
A data-driven method for predicting the \ac{RUL} of lithium-ion batteries involves analyzing the battery's operating data to estimate its degradation level and predict when it will reach the end of its useful life. This method relies on the analysis of battery performance metrics such as voltage, current, and temperature, as well as historical data. The data-driven method can directly mine the degradation information of lithium-ion battery through historical data, and there is no need to establish a specific mathematical model \cite{16}. The data-driven \ac{RUL} prediction method is a widely used method at this stage \cite{17}. Overall, data-driven \ac{RUL} prediction methods have the potential to improve battery performance and reduce costs associated with premature battery failure, making them an important area of research and development for the energy storage industry. This section breaks down the data driven methods into Traditional Machine Learning and Deep Learning . These subcategories encompass various techniques and algorithms tailored to efficiently mine and leverage battery data for more accurate predictions. The combination of these approaches holds the promise of advancing battery reliability and reducing the impact of unexpected failures, further fueling progress in the energy storage industry.

\vspace{0.15 cm}
\subsubsection{\textbf{Traditional Machine Learning}}
\begin{figure}[htbp]
    \centering
    \includegraphics[width=\linewidth]{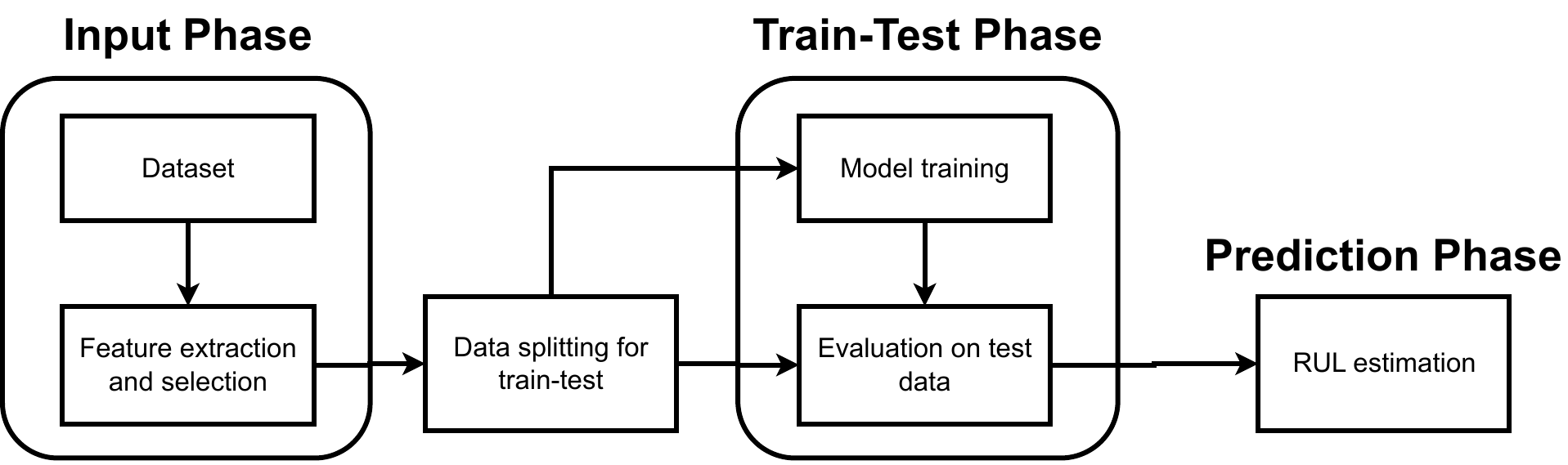}
    \caption{An overview of a process involving Machine Learning for predicting \ac{RUL}}
    \label{Machine Learning based RUL estimation process}
\end{figure}

\vspace{0.15 cm}
Machine learning methods have become favored choices, largely owing to their reliance on extensive historical data and their track record of delivering high accuracy. Fig. \ref{Machine Learning based RUL estimation process} illustrates the procedural framework for prognosticating Remaining Useful Life (RUL) through Machine Learning processes.

This section provides an overview of traditional machine learning methods for \ac{RUL} estimation of \ac{Li-ion} batteries. Zhou et al.\cite{19} proposed an incremental optimized\ac{RVM} that achieves efficient online training for model updating using battery data from the \ac{NASA} \ac{PCoE}.While this method can provide accurate predictions, it requires a large amount of training data and the selection of hyper parameters can be challenging. Refer \cite{19} for mathematical computations of the \ac{RVM} output .
In reference, Qin et al.\cite{20} proposed a \ac{FVS} and \ac{RVM} approach that addresses the memory consumption issue. The method involves using \ac{RVM} to estimate the \ac{RUL} and \ac{FVS} to remove redundant data points from the input data . However, the feature selection process may introduce bias or overlook important features, leading to sub-optimal performance. Additionally, Chen et al. \cite{21} uses \ac{CS} based \ac{RVM} with hybrid kernel for increased accuracy.This approach uses the \ac{CS} algorithm to intelligently determine kernel parameters and their weights.While this method shows promising results, the complexity of the kernel functions and their parameters can be a concern.
To further improve prediction accuracy, multiple ensemble models have been proposed. Li et al.\cite{22} proposed \ac{RUL} estimation based on Ensemble Learning with 
\ac{LS-SVM}, while Patil et al.\cite{23} proposed a multistage \ac{SVM} based approach that integrates the classification and regression attributes of \ac{SV} based machine learning technique. Wang et al. \cite{24} proposed \ac{RUL} estimation using \ac{SVR} optimized by \ac{ABC}. The \ac{ABC} algorithm optimizes \ac{SVR} core parameters.
Over the years, traditional machine learning methods have been widely used for \ac{RUL} estimation of \ac{Li-ion} batteries.Overall, these methods offer a range of approaches to address the challenge of \ac{RUL} estimation and can be useful in various applications, but careful consideration is needed when selecting the most appropriate method for a particular scenario.

   \ac{ML} based prediction process can be represented as:

\vspace{0.15 cm}
\begin{equation}
    \hat{Y}^* = M(X^*; p_{\text{opt}}
)
\end{equation}

Where:
\begin{itemize}
    \item $X^*$ is the new input data after pre-processing (including feature extraction and selection).
    \item $M$ is the trained machine learning model.
    \item $p_{\text{opt}}$ represents the optimized model parameters.
    \item $\hat{Y}^*$ is the predicted output.
\end{itemize}

\vspace{0.15 cm}
\subsubsection{\textbf{Gradual shift from Traditional ML to Deep Learning}}
\begin{figure}[htbp]
\centering
\begin{subfigure}[b]{\columnwidth}
        \centering
        \includegraphics[width=0.8\columnwidth]{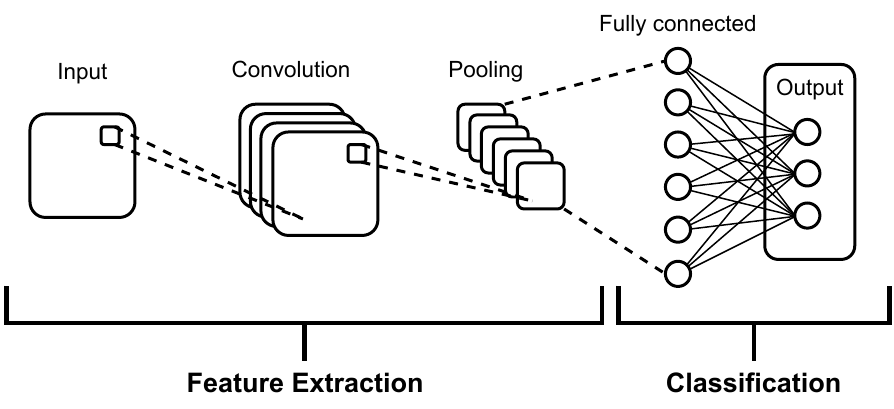}
   \caption{A high-level overview of a Convolutional Neural Network}
   \label{CNN}
\end{subfigure}
\\~\\
\begin{subfigure}[b]{\columnwidth}
\centering
   \includegraphics[width=0.8\columnwidth]{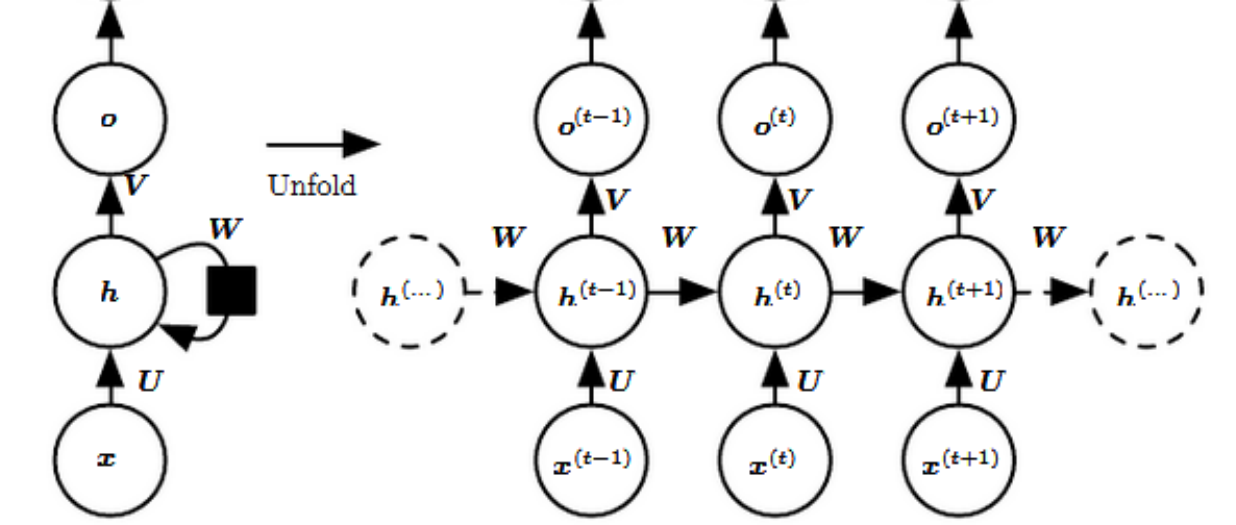}
   \caption{Unfolding of time in a Recurrent Neural Network  \cite{167}}
   \label{RNN}
\end{subfigure}
\\~\\
\begin{subfigure}[b]{\columnwidth}
\centering
   \includegraphics[width=0.9\columnwidth]{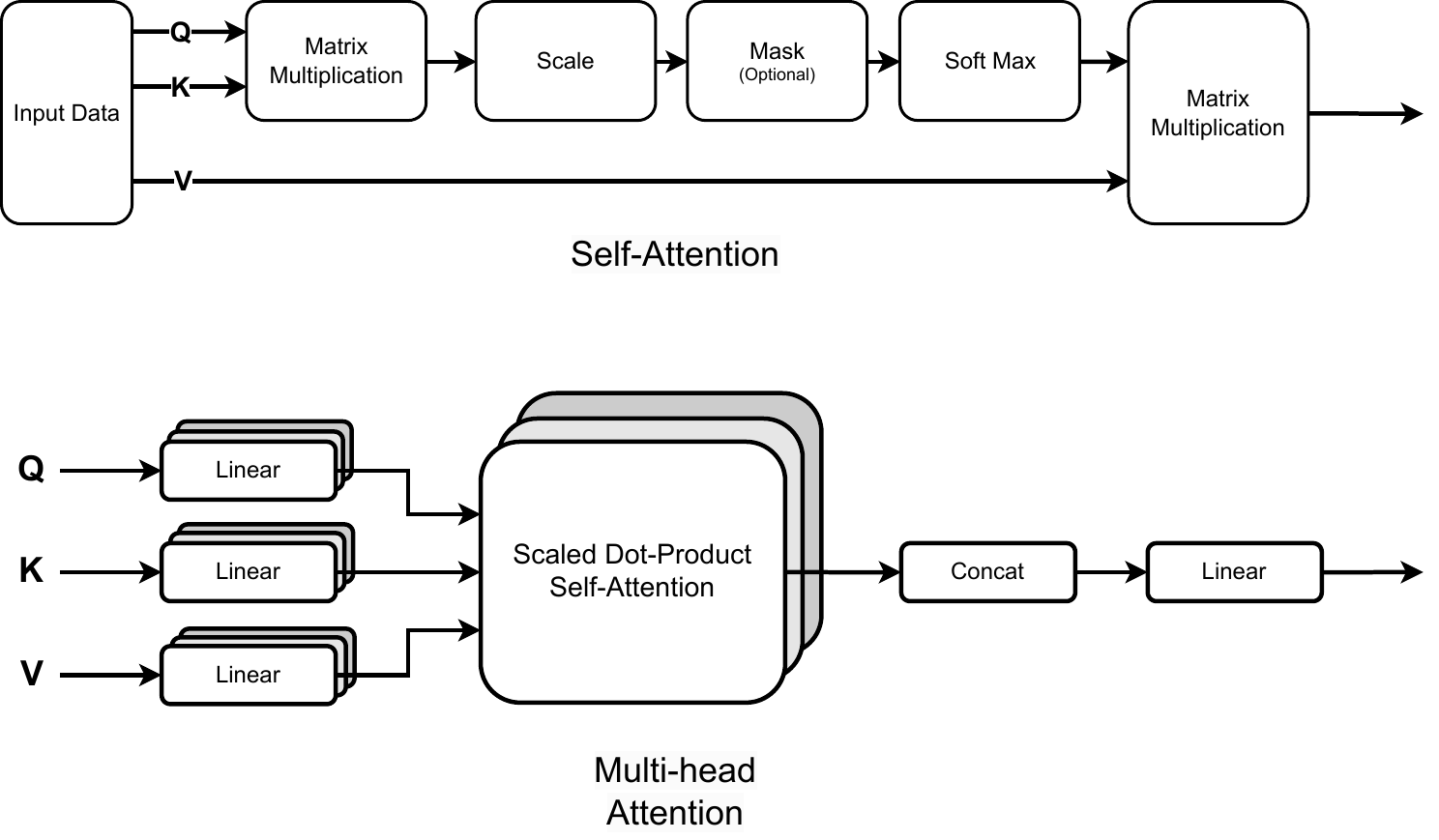}
   \caption{Self-attention and Multi-head attention modules employed in a Transformer \cite{90}}
   \label{Attention modules}
\end{subfigure}
\\~\\
\begin{subfigure}[b]{\columnwidth}
\centering
   \includegraphics[width=0.8\columnwidth]{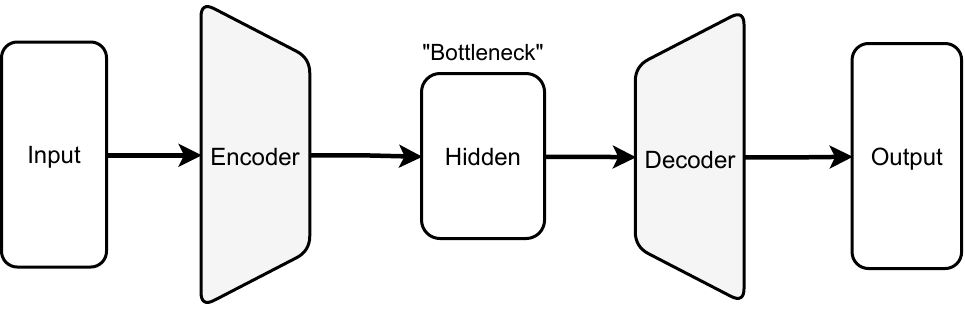}
   \caption{A schematic of an Autoencoder \cite{168}}
   \label{Deep Autoencoder}
\end{subfigure}
\caption{Schematic diagrams of various deep learning architectures}
\label{--}
\end{figure}

\vspace{0.15 cm}
The shift from traditional machine learning to deep learning is primarily motivated by the capacity of deep learning models to automatically extract intricate features from raw data, their superior performance in various applications, and increased accessibility to hardware and frameworks. This transition signifies a significant advancement in the field of machine learning, offering new and powerful tools for data analysis and pattern recognition. Some of the key points are briefed Refer \ref{Major Contributions in Deep Learning Research} for key advancements in this area.Refer \ref{CNN}, \ref{RNN} , \ref{Attention modules}, \ref{Deep Autoencoder} , for architecture diagrams.

\vspace{0.15 cm}
\begin{enumerate}
    \item \textbf{Feature Learning}: Deep learning models, such as neural networks, excel at feature learning, enabling them to automatically extract complex and hierarchical features from data, reducing the need for handcrafted feature engineering in traditional machine learning.
    \item \textbf{Performance Gains}: Deep learning has demonstrated remarkable performance improvements in tasks like image recognition, speech processing, and natural language understanding, outperforming traditional methods in many benchmark data sets
    \item \textbf{Transfer Learning}: Deep learning allows for effective transfer learning, where pre-trained models can be fine-tuned on specific tasks, reducing the need for extensive training on new data.
    \item \textbf{Scalability}: Deep learning models can be scaled to handle massive data sets and complex problems, making them suitable for the big data era and applications like deep reinforcement learning for autonomous systems.
    \item \textbf{Accessibility}:The availability of open-source deep learning frameworks like TensorFlow, PyTorch and increased computational resources has made deep learning more accessible to researchers and practitioners.
    \item \textbf{Versatality}:Deep learning can be applied across various domains, from healthcare and finance to autonomous vehicles, making it a versatile tool for a wide range of real-world applications.
    \item \textbf{Continuous Advancements}: Ongoing research and innovations in deep learning techniques ensure that the field remains dynamic, with new models and algorithms constantly emerging to address complex challenges.
\end{enumerate}

\begin{table}[tbh]
\centering
\renewcommand{\arraystretch}{2}
\scalebox{0.75}{%
  \begin{tabular}{|p{2.2cm}|p{7cm}|}
    \hline
    \multicolumn{1}{|c|}{\textbf{Author(s)}} & \multicolumn{1}{c|}{\textbf{Contributions}} \\
    \hline
    \specialrule{0.1em}{0.05em}{0.05em}
    \multirow{2}{*}{\centering LeCun et al.} & Pioneered Convolutional Neural Networks for image \newline recognition and analysis \cite{86} \\
    \hline
     \multirow{2}{*}{\centering Hopfield} & Introduced a special case of Recurrent Neural Networks for \newline sequential data processing (1982) \cite{87}\\
    \hline
    \multirow{2}{*}{\centering Rumelhart et al.} & Autoencoders to learn internal representations by error \newline propagation (1986) \cite{93} \\
    \hline
    \multirow{2}{*}{\centering Hochreiter et al.} & Addressed the problem of vanishing gradients in RNN with \newline Long Short Term Memory network (1997) \cite{88} \\
    \hline
    \multirow{1}{*}{\centering Cho et al.} & Gated Recurrent Units (2014)\cite{89} \\
    \hline
    \multirow{2}{*}{\centering Goodfellow et al.} & Generating realistic synthetic data with Generative \newline Adversarial Nets (2014) \cite{91} \\
    \hline
     \multirow{1}{*}{\centering Gatys et al.} & Neural Style Transfer (2015) \cite{94} \\
    \hline
    \multirow{2}{*}{\centering Vaswani et al.} & Enhancing context understanding in models with \newline 
    attention mechanism (2017) \cite{90} \\
    \hline
    \multirow{2}{*}{\centering Delvin et al.} & Bidirectional Encoder Representations from Transformers (2019) \cite{92} \\
    \hline
  \end{tabular}
}%
\caption{Major Contributions in Deep Learning Research}
\label{Major Contributions in Deep Learning Research}
\end{table}

\vspace{0.15 cm}
\subsubsection{Importance of Data Denoising}
\vspace{0.15 cm}
Data denoising is a critical pre-processing step in various data-driven applications, and it holds particular significance in the domain of Remaining Useful Life (RUL) estimation tasks. It involves the removal of unwanted or irrelevant noise from raw data to reveal the underlying patterns and trends. In the context of RUL estimation, which aims to predict the time until a machine or system's failure, accurate data is paramount for reliable predictions.

One widely employed technique for data denoising is the use of auto-encoders. Autoencoders are neural network architectures designed to encode data into a lower-dimensional representation and then decode it back to its original form. During this process, the network learns to filter out noise and capture the essential features of the data. Other denoising techniques include filtering methods such as median or moving average filters, and more advanced methods like principal component analysis (PCA) \cite{175}. These methods work by emphasizing signal components and attenuating noise.

The importance of data denoising in RUL estimation tasks cannot be overstated. Accurate RUL predictions rely on high-quality input data, as noise and anomalies can lead to incorrect estimations. Noise can obscure degradation patterns or introduce false alarms, which can be costly in real-world industrial applications. Denoising not only improves prediction accuracy but also enhances the interpretability of models, enabling better insights into the machinery's health. Thus, data denoising serves as a fundamental step in ensuring the success of RUL estimation models in various industries.

\vspace{0.15 cm}
\subsubsection{\textbf{Deep Learning}}
\vspace{0.15 cm}
Deep learning is a type of machine learning that uses neural networks to process and learn from data.

Various Deep Learning techniques have been proposed for \ac{RUL} estimation, including the \ac{LSTM}-\ac{RNN} method proposed by Zhang et al.\cite{25}, which captures long-term temporal dependencies of capacity degradation by using a sequence of memory cells to store past information. Ren et al. \cite{26} proposed an integrated Deep Learning approach that combines the Autoencoder with \ac{DNN} to extract multi-dimensional features and estimate \ac{RUL}. Liu et al. \cite{27} proposed a \ac{BMA} and \ac{LSTM} ensemble method that accounts for uncertainty and provides better prediction accuracy. Zhang et al. \cite{28} proposed an online estimation method that combines partial incremental capacity with an \ac{ANN} for estimating battery \ac{SOH} and \ac{RUL}. Zhou et al. \cite{29} proposed a \ac{TCN} based framework that uses causal and dilated convolution techniques to capture local capacity regeneration and improve prediction accuracy. Ren et al.\cite{30} proposed an Autoenocoder-Convolutional Neural Network-Long Short Term Memory Model that mines deeper information in finite data by using an Autoencoder to extract features from raw data and a combination of convolutional and \ac{LSTM} layers to learn temporal patterns in the data.Additionally, Li et al. \cite{31} designed an AST-LSTM NN for multiple battery sharing predictions, which uses a variant of the \ac{LSTM} model to predict the \ac{RUL} of multiple batteries.

In summary, Deep Learning techniques have shown promise in enhancing \ac{RUL} estimation for diverse applications in battery health management.

\begin{figure*}[htbp]
\centering
\begin{subfigure}[t]{0.6\textwidth}
        \centering
        \includegraphics[width=0.8\textwidth]{NASA_Battery.pdf}
   \caption{Baseline model predictions on NASA Dataset \cite{169}}
   \label{baseline-nasa}
\end{subfigure}
~
\begin{subfigure}[t]{0.3\textwidth}
\centering
   \includegraphics[width=0.8\textwidth]{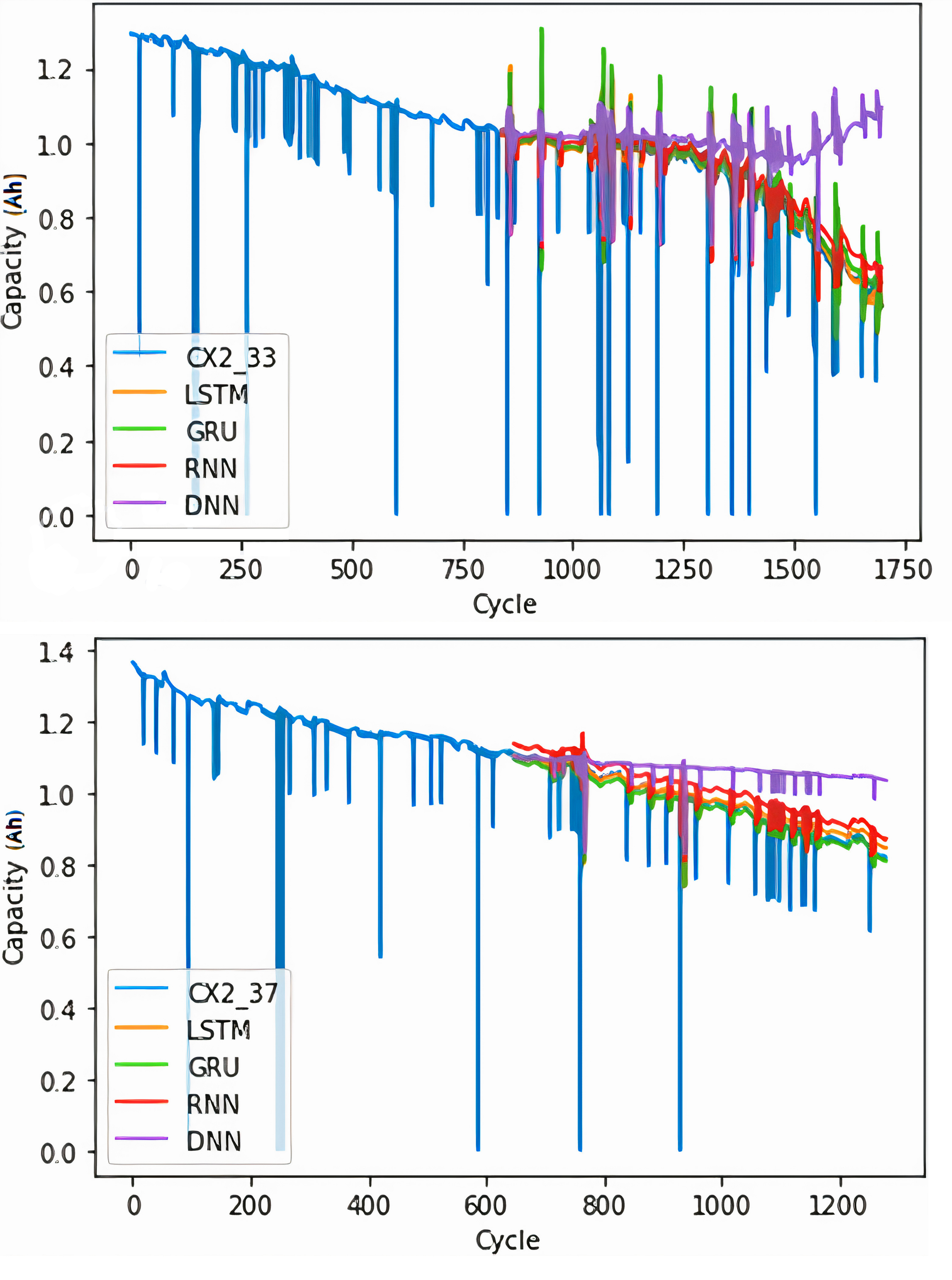}
   \caption{Baseline model predictions on CALCE Dataset \cite{169}}
   \label{CALCE}
\end{subfigure}
\caption{RUL estimation on NASA and CALCE data by various deep learning architectures}
\label{baseline-calce}

\end{figure*}

Below are some foundational deep learning architectures, including Long Short-Term Memory (LSTM), Gated Recurrent Unit (GRU), Recurrent Neural Networks (RNN), and Deep Neural Networks (DNN). Figures \ref{baseline-nasa} and \ref{baseline-calce} \cite{161} visually illustrate their data modeling techniques for Remaining Useful Life (RUL) estimation. These selections highlight the idea that using computationally efficient architectures can still deliver good performance. \cite{160} demonstrates the impressive performance of the complex Transformer architecture, showcasing notably low error metrics such as RE, MAE, and RMSE.Moreover, the figure also illustrates the enhanced capabilities of the De-Transformer, which combines the Transformer encoder with a Denoising Autoencoder. This helps in achieving better predictions by introducing uncertainty in the input data and learning representations from the same. \cite{162} further extends this concept by introducing different noise types, such as Gaussian, Uniform, and Speckle, through a comprehensive denoising framework.This framework combines a wavelet denoiser with a denoising autoencoder, effectively integrating them into a unified architecture.The model's output, represented by the lowest error metric marks a significant step forward in the research area.

\begin{figure*}[t!]
  \centering
      \includegraphics[width=\linewidth]{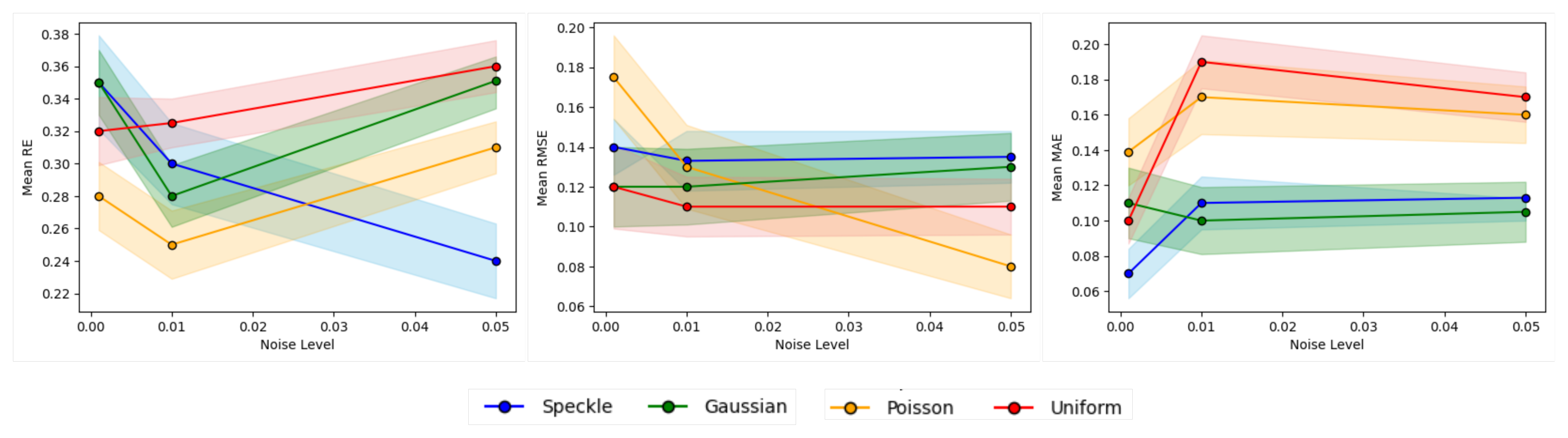}
      \caption{\textbf{95\%} Confidence interval for mean RE, RMSE, MAE for De-SaTE across different noise types on NASA data}
    \label{1}
      \vspace*{-0.1in}
\end{figure*}

\begin{figure*}[t!]
  \centering
      \includegraphics[width=\linewidth]{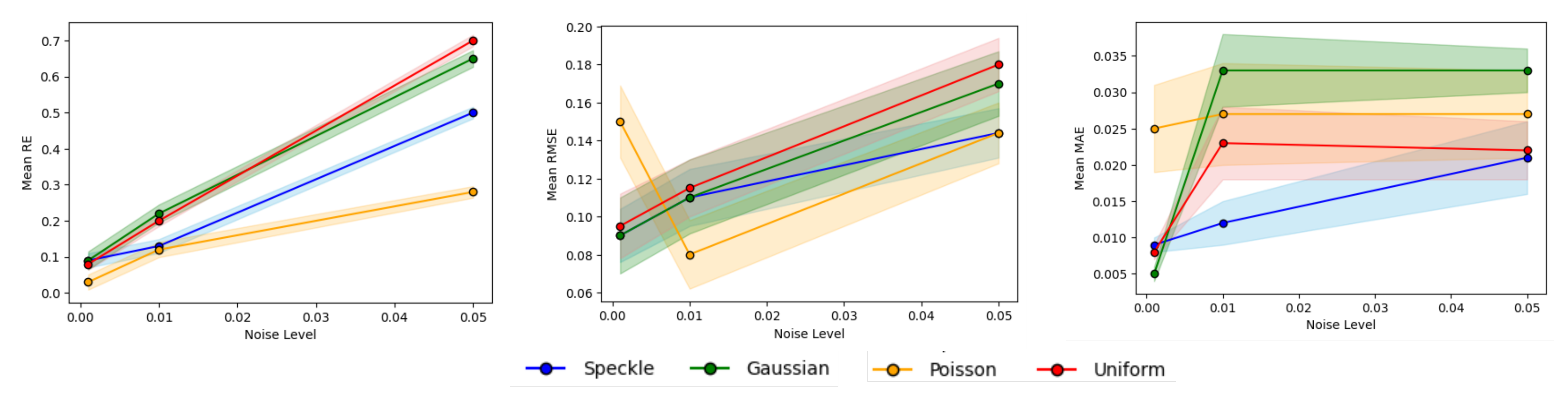}
      \caption{\textbf{95\%} Confidence interval for mean RE, RMSE, MAE for De-SaTE across different noise types on CALCE data}
    \label{2}
      \vspace*{-0.1in}
\end{figure*}

\begin{table}[htbp]
\centering
\caption{Model Evaluation on \ac{Li-ion} Battery Datasets}
\label{Model evaluation}
\scalebox{1}{\begin{tabular}{|l|l|ccc|}
\hline
Dataset & \multicolumn{1}{c|}{Model} & \multicolumn{3}{c|}{Metrics} \\ 
\hline
\specialrule{0.1em}{0.05em}{0.05em}
\textbf{} & \multicolumn{1}{|c|}{\textbf{}} & \multicolumn{1}{c}{\textbf{RE}} & \multicolumn{1}{c}{\textbf{MAE}} & \textbf{RMSE} \\ \hline
\multicolumn{1}{|c|}{\multirow{6}{*}{\textbf{NASA}}} & MLP \cite{MLP} & \multicolumn{1}{c}{0.3851} & \multicolumn{1}{c}{0.1379} & 0.1541 \\ 
\multicolumn{1}{|c|}{} & RNN \cite{RNN} & \multicolumn{1}{c}{0.2851} & \multicolumn{1}{c}{0.0749} & 0.0848 \\ 
\multicolumn{1}{|c|}{} & LSTM \cite{LSTM} & \multicolumn{1}{c}{0.2648} & \multicolumn{1}{c}{0.0829} & 0.0905 \\ 
\multicolumn{1}{|c|}{} & GRU \cite{GRU} & \multicolumn{1}{c}{0.3044} & \multicolumn{1}{c}{0.0806} & 0.0921 \\ 
\multicolumn{1}{|c|}{} & Dual-LSTM \cite{DUAL-LSTM} & \multicolumn{1}{c}{0.2557} & \multicolumn{1}{c}{0.0815} & 0.0879 \\
\multicolumn{1}{|c|}{} & Transformer \cite{DUAL-LSTM} & \multicolumn{1}{|c}{0.2700} & \multicolumn{1}{c}{0.1100} & 0.1300\\
\multicolumn{1}{|c|}{} & DeTransformer \cite{57} & \multicolumn{1}{c}{0.2252} & \multicolumn{1}{c}{\textbf{0.0713}} & 0.0802 \\
\multicolumn{1}{|c|}{} &\textbf{De-SaTE} \cite{169} & \multicolumn{1}{c}{\textbf{0.1674}} & \multicolumn{1}{c}{0.0806} & \textbf{0.0781} \\
\hline
\specialrule{0.1em}{0.05em}{0.05em}
\multirow{6}{*}{\textbf{CALCE}} & MLP \cite{MLP} & \multicolumn{1}{c}{0.4018} & \multicolumn{1}{c}{0.1557} & \multicolumn{1}{c|}{0.2038} \\ 
\multicolumn{1}{|c|}{} & RNN \cite{RNN} & \multicolumn{1}{c}{0.1614} & \multicolumn{1}{c}{0.0938} & 0.1099\\ 
\multicolumn{1}{|c|}{} & LSTM \cite{LSTM} & \multicolumn{1}{c}{0.0902} & \multicolumn{1}{c}{0.0582} & 0.0736 \\ 
\multicolumn{1}{|c|}{} & GRU \cite{GRU} & \multicolumn{1}{c}{0.1319} & \multicolumn{1}{c}{0.0671} & 0.0946 \\ 
\multicolumn{1}{|c|}{} & Dual-LSTM \cite{DUAL-LSTM} & \multicolumn{1}{c}{0.0885} & \multicolumn{1}{c}{0.0636} & 0.0874 \\
\multicolumn{1}{|c|}{} & Transformer \cite{DUAL-LSTM} & \multicolumn{1}{c}{0.1200} & \multicolumn{1}{c}{0.1000} & 0.125\\
\multicolumn{1}{|c|}{} & DeTransformer \cite{57} & \multicolumn{1}{c}{0.0764} & \multicolumn{1}{c}{0.0613} & \textbf{0.0705} \\
\multicolumn{1}{|c|}{} & \textbf{De-SaTE} \cite{169} & \multicolumn{1}{c}{\textbf{0.0330}} & \multicolumn{1}{c}{\textbf{0.0080}} & 0.090 \\
\hline
\specialrule{0.1em}{0.05em}{0.05em}
\end{tabular}}
\end{table}
\vspace{0.15 cm}

\vspace{0.15 cm}
\begin{figure}[htbp]
    \centering
    \includegraphics[width=\linewidth]{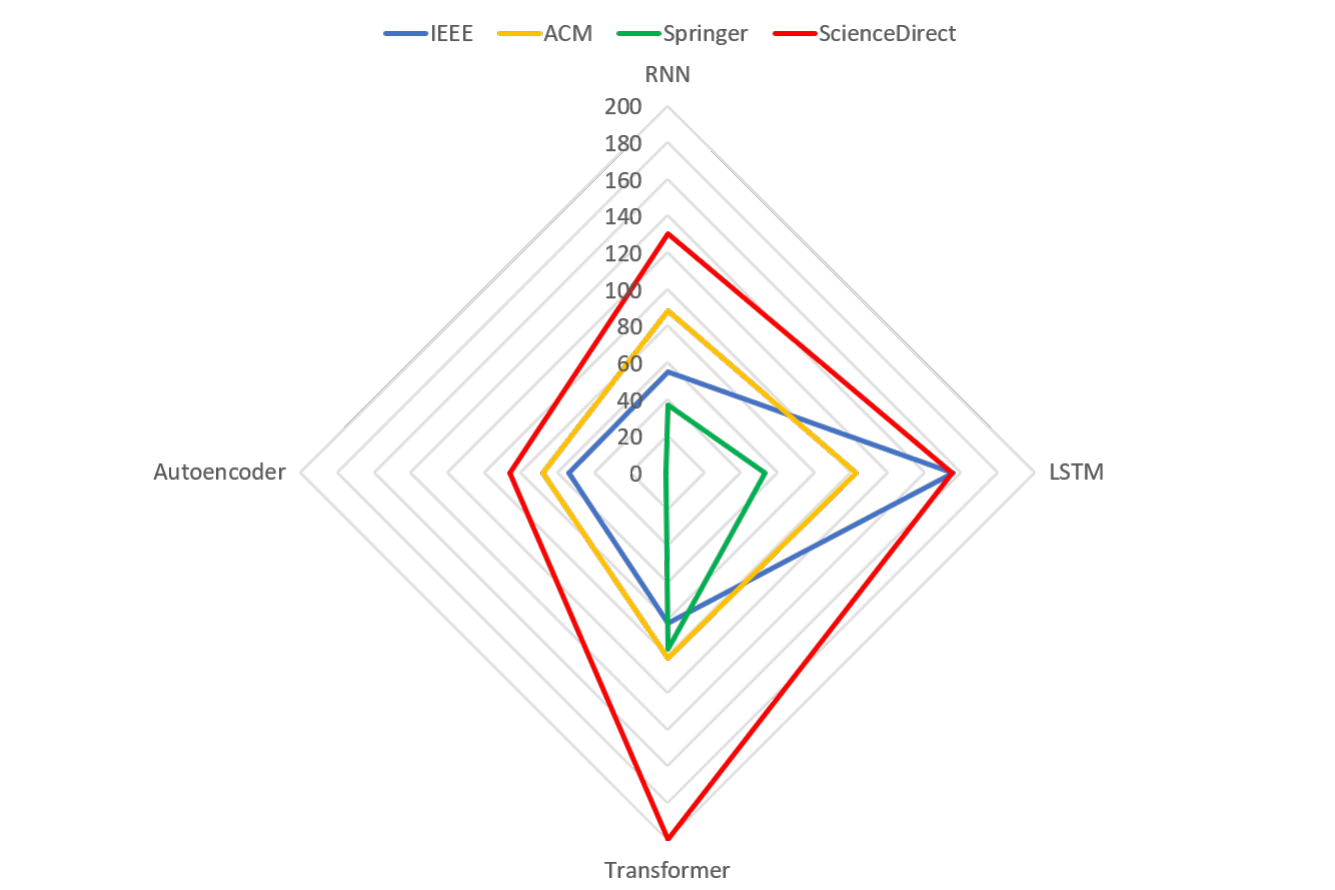}
    \caption{Deep Learning architectures for \ac{RUL} estimation of Li-ion batteries published 2021 and onwards}
    \label{Radar chart}
\end{figure}

\vspace{0.15 cm}
\section{Industrial Applications}
\vspace{0.15 cm}
\subsection{\textbf{Transportation Sector}}
\vspace{0.15 cm}

In the realm of the Transportation Sector, precise \ac{RUL} estimation is of paramount importance, with a particular focus on lithium-ion batteries that power various modes of transportation. The sector includes electric vehicles,  and electric aircraft, all relying on lithium-ion batteries for their energy storage and propulsion. Effective \ac{RUL} estimation for these batteries is essential for operational efficiency, safety, and cost-effectiveness. Lithium-ion batteries are mission-critical components, and accurately predicting their \ac{RUL} allows for proactive maintenance, optimal resource allocation, and cost reduction. Deep learning-based \ac{RUL} estimation models, rooted in data-driven methodologies, offer significant advantages for this sector \cite{95}.They empower operators to foresee when lithium-ion batteries are likely to reach the end of their operational life, enabling timely replacements and minimizing the risk of unexpected failures\cite{96}. This, in turn, ensures uninterrupted service, enhances passenger and cargo safety, and minimizes downtime, all while adhering to the unique challenges posed by lithium-ion battery systems.Moreover, data-driven decision-making, facilitated by advanced \ac{RUL} estimation models, contributes to the overall sustainability and competitiveness of the Transportation Sector.In a landscape marked by rapid technological advancements and evolving regulatory standards, \ac{RUL} estimation for lithium-ion batteries plays a central role in the sector's drive to enhance efficiency, elevate safety standards, and ensure fiscal prudence. In this context, the focus on lithium-ion batteries underscores the necessity for precision, reliability, and the elimination of redundancy in maintenance and operational practices, ultimately leading to a safer, more cost-effective, and sustainable future for transportation systems.

\vspace{0.15 cm}
\subsection{\textbf{Consumer Electronics}}
\vspace{0.15 cm}
In the domain of Consumer Electronics, precise \ac{RUL} estimation for lithium-ion batteries holds a pivotal role in guaranteeing the seamless operation of personal devices.The last 15 years have seen enormous changes in consumer electronics technologies, and a great expansion in the variety of devices available at relatively low cost.Entertainment, communication, and computing technologies have tended to merge, and almost everything is 'going digital'\cite{97}. Accurate Remaining Useful Life (\ac{RUL}) estimation for lithium-ion batteries is pivotal in ensuring the uninterrupted operation of these devices, which range from smartphones and laptops to wearable technology, all of which rely on lithium-ion batteries to power everything from communication to productivity.In these days, the portable electronics are fully-featured than ever before and the users are increasingly dependent on these mobile devices \cite{98} and thus accurate \ac{RUL} prediction for these batteries is a linchpin for uninterrupted device functionality, user safety, and cost-effective ownership. Advanced \ac{RUL} estimation models, particularly those founded on data-driven methodologies and deep learning, offer a host of benefits. These models empower both consumers and manufacturers to anticipate when lithium-ion batteries are nearing the end of their operational life, enabling timely replacements and mitigating the risk of unexpected battery depletion. Furthermore, the incorporation of data-informed decision-making, facilitated by these state-of-the-art models, contributes to the sector's sustainability and competitive edge.In an ever-evolving landscape where consumer expectations are on the rise, \ac{RUL} estimation for lithium-ion batteries takes center stage in the Consumer Electronics sector's commitment to delivering reliable, efficient, and cost-effective devices.

\vspace{0.15 cm}
\subsection{\textbf{Healthcare}}
\vspace{0.15 cm}
In the Healthcare Sector, the landscape has undergone a transformative evolution over the past two decades.The advent of advanced medical technologies has led to a proliferation of portable medical devices, monitoring equipment, and implantable sensors, heralding a new era in patient care and diagnostics.Devices used in healthcare applications frequently require powering\cite{99}.These healthcare devices usually comprise rechargeable and portable \ac{Li-ion} batteries\cite{100}.These batteries power portable defibrillators, insulin pumps and remote patient monitors.Accurate \ac{RUL} estimation for these batteries is fundamental to ensuring the reliability and uninterrupted function of these life-saving tools.Cutting-edge \ac{RUL} estimation models, particularly those rooted in data-driven methodologies and deep learning, have ushered in a new paradigm of predictive maintenance. These models empower healthcare providers to foresee when lithium-ion batteries are nearing the end of their operational life, enabling timely replacements and mitigating the risk of battery failure. This proactive approach ensures the continuous and precise functionality of critical medical devices, enhances patient safety, and minimizes disruptions in healthcare delivery. 

\vspace{0.15 cm}
\section{Battery Health and Cyber-Physical Systems}
The meticulous exploration of \ac{RUL} estimation for \ac{Li-ion} batteries in key sectors highlights a pivotal aspect of modern industrial activities: the dependence on advanced energy storage systems to ensure reliability and efficiency in an increasingly digital landscape. As we transition from the specific challenges and progress in battery prognostics to the wider industrial scene marked by \ac{CPS}, it becomes clear that these discussions are interconnected. The operational integrity and resilience of \ac{CPS}, which underpin critical infrastructure, rely not just on the batteries that power these systems but equally on the strength of the digital frameworks that protect them. In this vein, examining \ac{CPS} emerges as an integral part of our discourse on battery health prognostics, seamlessly connecting the dots between ensuring the physical durability of vital devices and securing the digital environments they function within. We also offer insights into industrial trends to ensure our readers stay informed about the latest developments.

\vspace{0.15 cm}
\section{CPS Insights}
\vspace{0.15 cm}
Whether in critical infrastructure, manufacturing, warehousing, transportation, utilities, building management or healthcare, every asset-intensive organization has Cyber-Physical Systems. They can be interchangeably called operational technology (OT), Internet of Things (IoT), Industrial IoT (IIoT), Internet of Medical Things (IoMT), smart building solutions, or Industrie 4.0 \cite{163}.Irrespective of the terminology organizations embrace, these systems share a common trait: they are digitally managed yet actively interface with the tangible physical world.

\vspace{0.15cm}
\subsection{Industry Scope}
\begin{figure}[htbp]
    \centering
    \includegraphics[width=0.95\columnwidth]{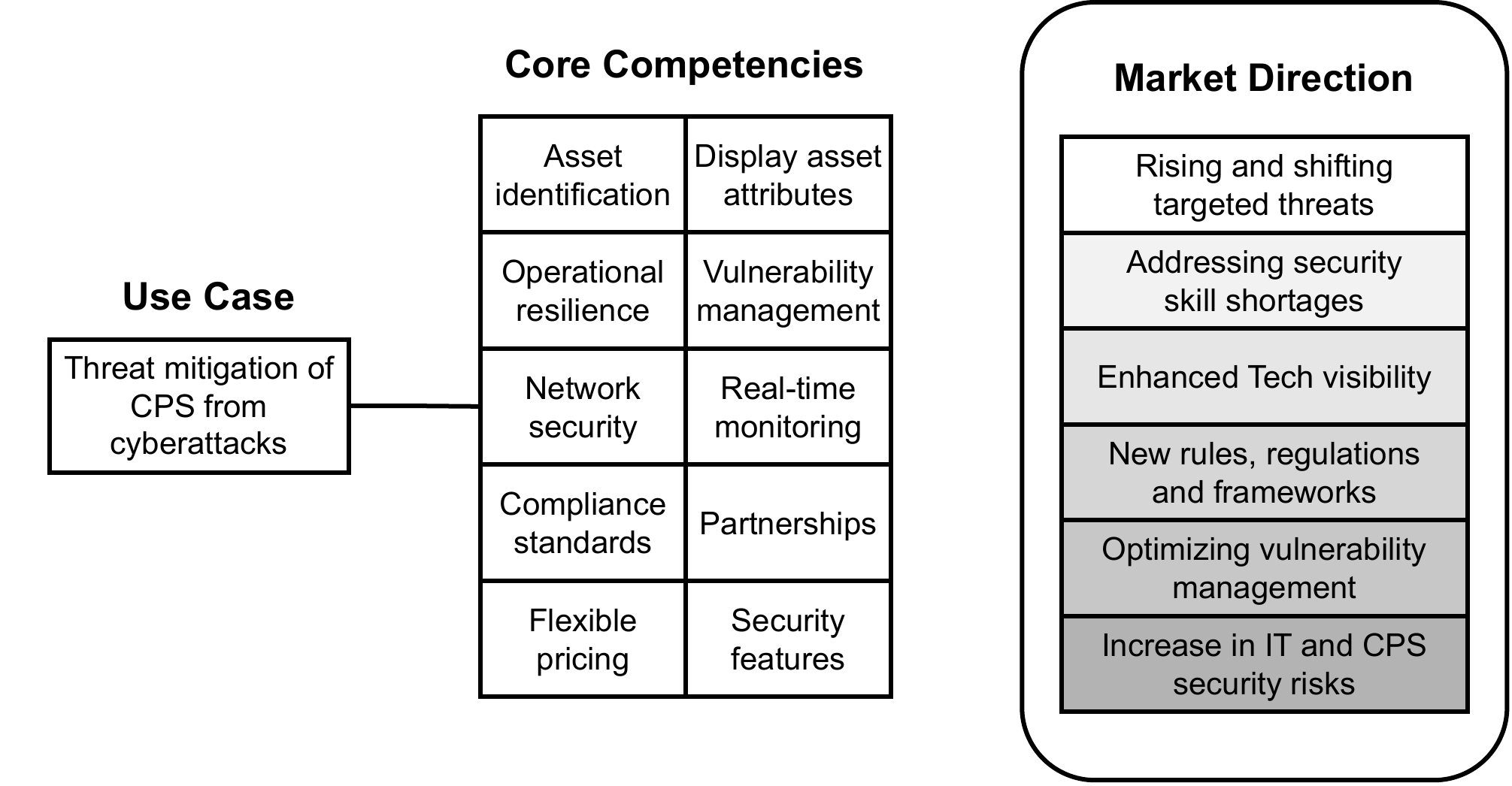}
    \caption{High-level overview of the CPS Protection Platform}
    \label{CPS protection platform overview}
\end{figure}

\vspace{0.15cm}
A leading tech blog defines the cyber-physical systems(CPS) protection platforms market as products and services that use knowledge of industrial protocols,operational/production network packets or traffic metadata, and physical process asset behavior to discover, categorize, map and protect CPS in production or mission-critical environments outside of enterprise IT environments \cite{163}.Figure \ref{CPS protection platform overview} shows an overview of CPS Protection Platforms.

\vspace{0.1 cm}
Key Platform Attributes :
\begin{itemize}
  \item Discovering and Categorizing CPS Assets
  \item Comprehensive Asset Pedigree
  \item Support for Industrial Protocols
  \item Clear Network Maps and Data Flow
  \item Integration with IT Security Tools
  \item Real-time Threat Detection
\end{itemize}

\vspace{0.15 cm}
\subsection{Emerging Domains within Industries 4.0 Due to Evolving Threats}
\vspace{0.15 cm}
The security of industrial cyber-physical systems (CPS) is paramount, as they face various threats,each posing unique risks to critical operations.The most commonly encountered threats are:
\vspace{0.15cm}
\begin{itemize}
  \item \textbf{Malware and Ransomware Attacks:} Malicious software, including viruses and ransomware, can disrupt CPS operations by infecting control systems, leading to data loss or demanding ransoms for system restoration.

  \item \textbf{Unauthorized Access and Intrusions:} Unauthorized individuals or entities gaining access to industrial networks can manipulate or sabotage critical systems, potentially causing significant damage.

  \item \textbf{Physical Attacks:} These encompass physical intrusions, such as tampering with sensors or hardware, which can lead to misleading data or the compromise of physical safety in industrial settings.
\end{itemize}

\vspace{0.1 cm}
Due to the dynamic and swiftly evolving threat landscape, novel categories of CPS security solutions have arisen \cite{163}:

\begin{itemize}
  \item CPS content disarm and reconstruction solutions
  \item CPS cyber risk quantification platforms
  \item CPS firmware/embedded systems/device security
  \item CPS onboard diagnostics solutions
  \item CPS secure remote access solutions
  \item CPS security services
  \item CPS supply chain security solutions
\end{itemize}

\vspace{0.15 cm}
\subsection{CPS Market Size and Analysis}
\begin{figure}[htbp]
    \centering
    \includegraphics[width=\linewidth]{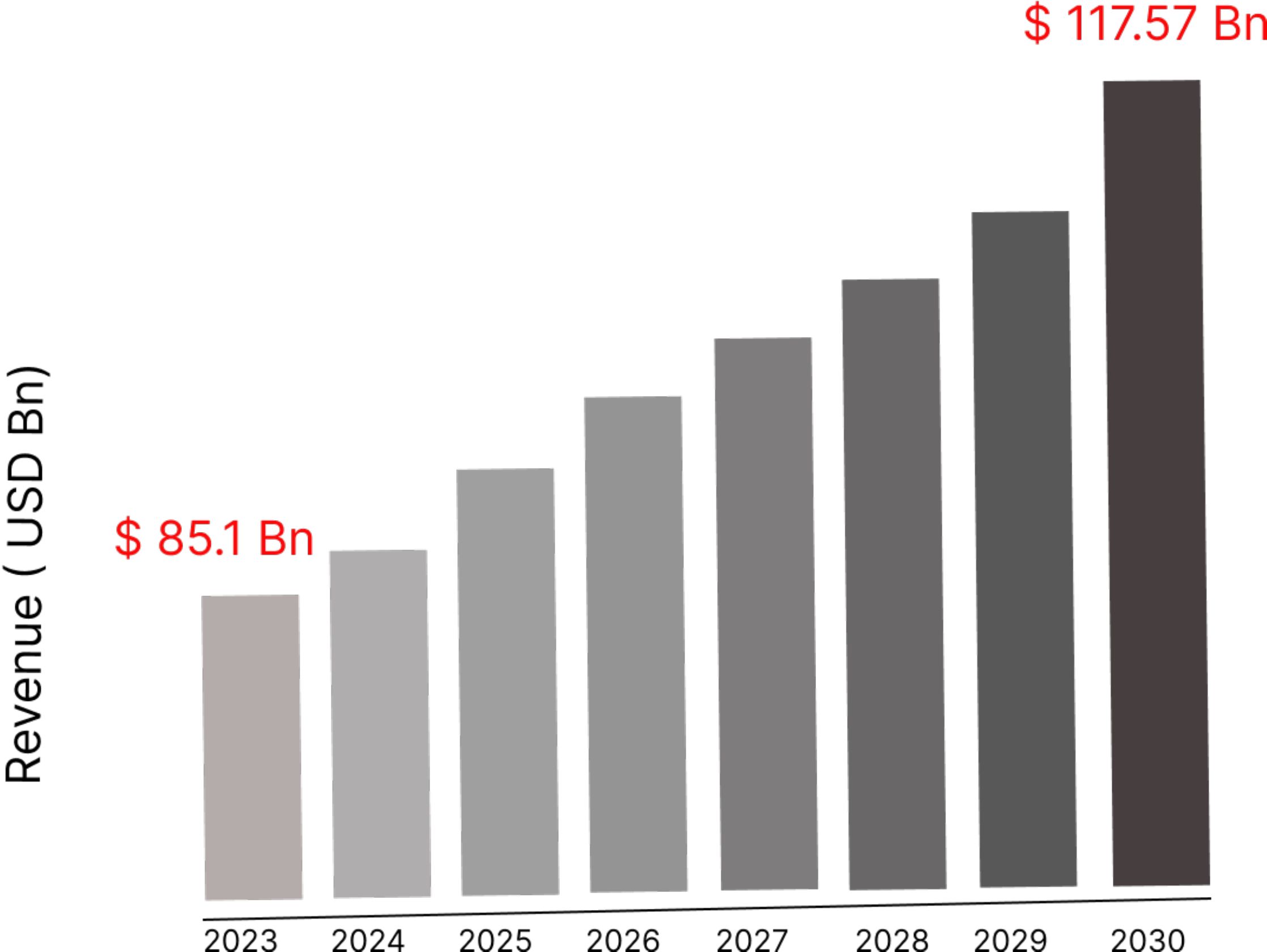}
    \caption{Projected CPS market size over the years}
    \label{CPS market trend}
\end{figure}

\vspace{0.15 cm}
The global cyber-physical systems (CPS) market size is predicted to increase from \$76.98 billion in 2021 to \$177.57 billion by 2030 at a CAGR of nearly 8.01\% during 2022-2030. \ref{CPS market trend} shows the drastic increase in revenue from 2018 to 2030 \cite{164}.

\vspace{0.1 cm}
Key insights: 
\begin{itemize}
    \item Based on component, The cyber-physical system hardware segment is anticipated to dominate a significant market share throughout the forecast period.
    \item On basis of vertical, the energy and utility segment is predicted to dominate the global market share over 2022-2030\cite{164}.
    \item When considering specific countries, United States and Germany are expected to emerge as the primary regional revenue hubs for the global cyber-physical systems (CPS) market between 2022 and 2030.
\end{itemize}

\vspace{0.15 cm}
\subsection{Competitive Landscape}
\vspace{0.15 cm}
The key players responsible for the expansion of Global Cyber-Physical System are Microsoft,Hewlett Packard Enterprise Development LP, Dell Inc, VMware Inc, Schneider Electric, SAP SE, Honeywell International Inc, Hitachi Vantara Corporation, Intel Corporation,Siemens, IBM Corporation, Oracle Corporation \cite{164}.

\vspace{0.15cm}
\subsection{Vendor Profiles}
Several leading vendors offer diverse solutions to protect and secure Cyber-Physical Systems (CPS).\ref{cps-vendors-table} explores a range of these solutions and their distinctive features.

\begin{table*}[h]
\centering
\renewcommand{\arraystretch}{2}
\caption{Key features and deployment options offered by various CPS security solution vendors}

\begin{tabular}{|p{2.1 cm}|p{9.2cm}|p{6.5cm}|}
\hline
\multicolumn{1}{|c|}{\textbf{\centering Vendor}} & \multicolumn{1}{|c|}{\textbf{\centering Key Features}} & \multicolumn{1}{|c|}{\textbf{\centering Deployment Options}} \\
\hline
\specialrule{0.1em}{0.05em}{0.05em}
AirEye & Asset discovery, classification, and connection data capture & Cloud-to-cloud integration, physical sensor deployment \\
\hline
Armis & Zero-touch collectors, traffic analysis, risk identification & On-premises sensors, cloud connections \\
\hline
Claroty & Multiple asset discovery methods, extensive protocol support & SaaS and on-premises deployment, sensor optimization \\
\hline
Cylera & Adaptive Data Type Analysis (ADTA), passive sensors, Io(M)T asset digital twin & Physical, containerized, and VM sensor deployment \\
\hline
Cylus & Network traffic data collection, asset classification, rail-specific focus & SaaS and on-premises deployment, sensor optimization \\
\hline
Darktrace & Self-learning AI, passive discovery, autonomous blocking & Physical appliances, SaaS, VMs \\
\hline
Dragos & Passive discovery, asset inventories, global support & Hardware, virtual, and cloud-based appliances \\
\hline
Forescout & Passive deep packet inspection, active queries, three-tier architecture & Docker container sensors, network switch integration \\
\hline
Fortinet & Fortinet Security Fabric, NGFW, FortiSwitch, FortiManager & On-premises, virtual, cloud-based deployments \\
\hline
Microsoft & Defender for IoT, network sensor, DPI, behavioral profiling & Self-service deployment via Microsoft Azure \\
\hline
Nozomi Networks & Passive network-based monitoring, endpoint sensors, endpoint asset details & On-premises sensors in various form factors \\
\hline
OPSWAT & Active and passive asset discovery, exposure score & Hardware, software, VM, on-premises or cloud-based \\
\hline
Ordr & Asset discovery via network traffic analysis and AI/ML algorithms & Hardware and software appliance sensors \\
\hline
OTORIO & Passive network monitoring, active queries, cyber digital twin & Physical or virtual appliances, on-premises or SaaS \\
\hline
Palo Alto Networks & Zero Trust OT Security Solution, NGFW, Cortex XSOAR & Hardware, software, and cloud service sensors \\
\hline
SCADAfence & Passive deep packet inspection, ecosystem enrichments & Software-based sensors, on-premises or cloud-based deployment \\
\hline
Tenable & Passive and active assessments, Nessus scanning & VM image, physical appliance, on-premises deployment \\
\hline
Verve Industrial & Discovery client, asset collection, ecosystem integrations & On-premises and cloud-based platform \\
\hline
Xage Security & Xage Fabric, access control, asset discovery, Xage Nodes & Overlay on existing infrastructure \\
\hline
\specialrule{0.1em}{0.05em}{0.05em}
\end{tabular}
\label{cps-vendors-table}
\end{table*}

\vspace{0.15cm}
\ref{Distribution of articles by Application Areas} offers an extensive compilation of literature sources, en-compassing various application areas for in-depth exploration.
\vspace{0.15 cm}

\begin{table}[tbh]
\centering
\renewcommand{\arraystretch}{2}
\scalebox{0.75}{%
  \begin{tabular}{|p{2.35cm}|p{7cm}|}
    \hline
    \multicolumn{1}{|c|}{\textbf{Application Area}} & \multicolumn{1}{c|}{\textbf{Authors}} \\
    \hline
    \specialrule{0.1em}{0.05em}{0.05em}
\multirow{4}{*}{\centering Transportation sector} & Dokgoz et al. \cite{68}, Zhang et al. \cite{69}, Omariba et al. \cite{66}, Wet et al. \cite{67}, Xing et al. \cite{101}, Sundararaju et al. \cite{102}, Shete et al. \cite{103}, Gao et al. \cite{104}, Livint et al. \cite{105}, Sun et al. \cite{106}, Samanta et al. \cite{107}, Amanathulla et al.\cite{108}, Wu et al. \cite{109}, Hannan et al. \cite{110}, Lin et al. \cite{111}, Xie et al. \cite{112}, Xiong et al. \cite{113}, Lee et al. \cite{114}\\
    \hline
     \multirow{2}{*}{\centering Consumer Electronics} & Ma et al. \cite{70}, Gokcen et al. \cite{71}, Jain et al. \cite{72}, Zhang et al. \cite{115}, Namdari et al. \cite{116}, Zhang et al. \cite{117}, Ilies et al. \cite{118}, Goud et al. \cite{119} \\
    \hline
     \multirow{2}{*}{\centering Healthcare} & Liu et al. \cite{79}, Hu et al. \cite{77}, Hu et al. \cite{78}, Makin et al. \cite{120}, Mohsen at al. \cite{121}, Spillman et al. \cite{122}, Pavlov et al. \cite{123}\\
     \hline
  \end{tabular}
}%
\caption{Distribution of articles by Application Areas}
\label{Distribution of articles by Application Areas}
\end{table}
\vspace{0.15 cm}

\begin{table*}[tbh]
\centering
\renewcommand{\arraystretch}{2}
\scalebox{0.75}{
  \begin{tabular}{|p{2cm}|p{16cm}|p{2.3 cm}|}
    \hline
    \multicolumn{1}{|c|}{\textbf{Area}} & \multicolumn{1}{c|}{\textbf{Paper Hyperlink}} & \multicolumn{1}{c|}{\textbf{Author}}
    \\
    \hline
    \specialrule{0.1em}{0.05em}{0.05em}
\multirow{20}{*}{\centering \textbf{Deep Learning}}
& \href{https://arxiv.org/abs/1906.05264}{GluonTS: Probabilistic Time Series Models in Python} \cite{146} & Alexandrov et al.
\\
\cline{2-3}
 & \href{https://journalofbigdata.springeropen.com/articles/10.1186/s40537-021-00444-8}{Review of deep learning: concepts, CNN architectures, challenges, applications, future directions} \cite{125} & Alzubaidi et al.
\\
   \cline{2-3}
   & \href{https://arxiv.org/abs/1406.2661v1}{Generative Adverserial networks} \cite{140} & Goodfellow et al.
\\
\cline{2-3}
   & \href{https://ieeexplore.ieee.org/document/7780459}{Deep Residual Learning for Image Recognition} \cite{138} & He et al.
\\
\cline{2-3}
& \href{https://www.sciencedirect.com/science/article/abs/pii/0893608089900208}{Multilayer feedforward networks are universal approximators} \cite{129} & Hornik et al.
\\
    \cline{2-3}
    & \href{https://www.sciencedirect.com/science/article/abs/pii/S0893608014002214}{Trends in extreme learning machines: A review} \cite{142} & Huang et al.
\\
\cline{2-3}
& \href{https://dl.acm.org/doi/10.5555/3045118.3045167}{Batch normalization: accelerating deep network training by reducing internal covariate shift} \cite{139} & Ioffe et al.
\\
\cline{2-3}
& \href{https://www.sciencedirect.com/science/article/pii/S0169207021000637}{Temporal Fusion Transformers for interpretable multi-horizon time series forecasting} \cite{145} & Lim et al.
\\
\cline{2-3}
& \href{https://dl.acm.org/doi/10.5555/3295222.3295230}{A unified approach to interpreting model predictions} \cite{149} & Lundberg et al.
\\
\cline{2-3}
 & \href{https://www.sciencedirect.com/science/article/pii/S089360802300014X}{A wholistic view of continual learning with deep neural networks: Forgotten lessons and the bridge to active and open world learning} \cite{130} & Mundt et al.
\\
    \cline{2-3}
     & \href{https://www.sciencedirect.com/science/article/pii/S0893608019300231}{Continual lifelong learning with neural networks: A review} \cite{127} & Parisi et al.
\\
    \cline{2-3}
    & \href{https://dl.acm.org/doi/10.1145/2939672.2939778}{"Why Should I Trust You?": Explaining the Predictions of Any Classifier} \cite{148} & Ribeiro et al.
\\
\cline{2-3}
& \href{https://arxiv.org/abs/2112.10752}{High-Resolution Image Synthesis with Latent Diffusion Models} \cite{143} & Rombach et al.
\\
\cline{2-3}
 & \href{https://link.springer.com/article/10.1007/s42979-021-00815-1}{Deep Learning: A Comprehensive Overview on Techniques, Taxonomy, Applications and Research Directions} \cite{126} & Sarker et al.
\\
    \cline{2-3}
 & \href{https://arxiv.org/abs/1905.13294}{A Review of Deep Learning with Special Emphasis on Architectures, Applications and Recent Trends} \cite{124} & Sengupta et al.
\\
    \cline{2-3}
    & \href{https://www.sciencedirect.com/science/article/abs/pii/S0893608014002135}{Deep learning in neural networks: An overview} \cite{128} & Schmidhuber et al.
\\
    \cline{2-3}
    & \href{https://arxiv.org/abs/1409.3215}{Sequence to Sequence Learning with Neural Networks} \cite{144} & Sutskever et al.
\\
\cline{2-3}

    & \href{https://www.sciencedirect.com/science/article/abs/pii/S0893608018303332}{Deep learning in spiking neural networks} \cite{131} & Tavnaei et al.
\\
    \cline{2-3}
& \href{https://arxiv.org/abs/1411.1792}{How transferable are features in deep neural networks?} \cite{141} & Yosinski et al.
\\
\cline{2-3}
& \href{https://arxiv.org/abs/2012.07436}{Informer: Beyond Efficient Transformer for Long Sequence Time-Series Forecasting} \cite{147} & Zhou et al.
\\
    \hline
     \specialrule{0.1em}{0.05em}{0.05em}
     \multirow{10}{*}{\centering\textbf{PHM} } 
     & \href{https://link.springer.com/article/10.1007/s13198-013-0195-0}{Remaining useful life estimation: review} \cite{135} & Ahmadzadeh et al.
\\
    \cline{2-3}
    & \href{https://doi.org/10.1214/aop/1176996548}{Residual Life Time at Great Age} \cite{159} & Balkema et al.
     \\
     \cline{2-3}
     & \href{https://link.springer.com/article/10.1007/s00184-008-0220-5}{Remaining useful life in theory and practice} \cite{158} & Banjevic et al.
     \\
     \cline{2-3}
      & \href{https://www.sciencedirect.com/science/article/pii/S0951832022000321}{Variational encoding approach for interpretable assessment of remaining useful life estimation} \cite{150} & Costa et al.
     \\
     \cline{2-3}
     & \href{https://www.sciencedirect.com/science/article/pii/S0278612522000796}{Remaining Useful Life prediction and challenges: A literature review on the use of Machine Learning Methods} \cite{133} & Fereirra et al.
\\
    \cline{2-3}
     & \href{https://www.researchgate.net/publication/366340099_Masked_Self-Supervision_for_Remaining_Useful_Lifetime_Prediction_in_Machine_Tools}{Masked Self-Supervision for Remaining Useful Lifetime Prediction in Machine Tools} \cite{153} & Guo et al.
     \\
     \cline{2-3}
       & \href{https://arxiv.org/abs/2212.14612}{Conformal Prediction Intervals for Remaining Useful Lifetime Estimation} \cite{151} & Javanmardi et al.
     \\
     \cline{2-3}
      & \href{https://ieeexplore.ieee.org/document/1548498}{Reasoning about uncertainty in prognosis: a confidence prediction neural network approach} \cite{156} & Khawaja et al.
     \\
     \cline{2-3}
      & \href{https://ieeexplore.ieee.org/abstract/document/7726039}{Direct Remaining Useful Life Estimation Based on Support Vector Regression} \cite{157} & Khelif et al.
     \\
     \cline{2-3}
       & \href{https://arxiv.org/abs/2108.08721}{Improving Semi-Supervised Learning for Remaining Useful Lifetime Estimation Through Self-Supervision} \cite{152} & Krokotsch et al. 
     \\
     \cline{2-3}
     & \href{https://journals.sagepub.com/doi/10.1177/0142331208092030}{Comparison of prognostic algorithms for estimating remaining useful life of batteries} \cite{136} & Saha et al.
\\
    \cline{2-3}
    & \href{https://www.sciencedirect.com/science/article/pii/S2590005623000462}{Learning remaining useful life with incomplete health information: A case study on battery deterioration assessment} \cite{154} & Sanchez et al.
     \\
     \cline{2-3}
    & \href{https://papers.phmsociety.org/index.php/phmconf/article/view/2263}{WHY IS THE REMAINING USEFUL LIFE PREDICTION UNCERTAIN?} \cite{134} & Sankararaman et al.
\\
    \cline{2-3}
    & \href{https://ideas.repec.org/a/eee/ejores/v213y2011i1p1-14.html}{Remaining useful life estimation - A review on the statistical data driven approaches} \cite{137} & Si et al.
     \\
     \cline{2-3}
      & \href{https://link.springer.com/article/10.1007/s10845-009-0356-9}{An artificial neural network method for remaining useful life prediction of equipment subject to condition monitoring} \cite{155} & Tian et al.
     \\
     \cline{2-3}
     & \href{https://www.sciencedirect.com/science/article/pii/S2212827114001140}{Overview of Remaining Useful Life Prediction Techniques in Through-life Engineering Services} \cite{132} & Okoh et al.
\\
    \hline
  \end{tabular}
}
\caption{Collection of Useful Papers}
\label{Collection of Useful Papers}
\end{table*}

\section{Conclusion and Future Work}

\vspace{0.15 cm}
In conclusion, we shed light on several avenues of research while delving into ongoing efforts and insights directed at tackling the multifaceted challenges that lie ahead.
\vspace{0.15 cm}
\begin{enumerate}
    \item \textbf{Addressing Data Scarcity in \ac{RUL} Estimation:} The challenge of data scarcity in \ac{RUL} estimation for lithium-ion batteries remains a pressing issue.Future research should explore novel data collection methodologies, including distributed sensors and data augmentation techniques, to address this concern.By increasing the volume and quality of available data, we can enhance the accuracy and reliability of \ac{RUL} predictions.
    \item \textbf{Advancements in Prognostics Techniques:} The field of battery prognostics is poised for remarkable advancements.Research should continue to refine and expand the toolkit of prognostic techniques, integrating machine learning, deep learning, and \ac{AI} based methods. Collaborations between domain experts and data scientists will be essential to create more accurate and adaptable models for \ac{RUL} estimation.
    \item \textbf{Incorporating Unsupervised Learning:} Unsupervised learning techniques hold immense potential for refining battery prognostics by allowing models to extract valuable insights from unlabeled data.As this area gains more attention, further research is needed to develop and adapt unsupervised learning approaches specifically for this task.These efforts will enhance the ability to draw predictive insights from diverse and unstructured data sources.
    \item \textbf{Mitigating Risks from Adversarial Attacks:} As battery systems become more integral to critical applications, ensuring their security is paramount.Future research should focus on recognizing and mitigating potential adversarial attacks that could compromise battery health assessments.Developing robust models and innovative defense strategies is vital to protect against security threats and maintain the trustworthiness of \ac{RUL} predictions.
\end{enumerate}

\vspace{0.15 cm}

\section{Abbreviations}
\begin{acronym}[AAAAA]
    \acro{Li-ion}{Lithium-ion}
    \acro{PHM}{Prognostics and Health Management}
    \acro{RUL}{Remaining Useful Life}
    \acro{KF}{Kalman Filter}
    \acro{RVM}{Relevance Vector Machine}
    \acro{UKF}{Unscented Kalman filter}
    \acro{AUKF}{Adaptive Unscented Kalman Filter}
    \acro{EOS}{End of Service}
    \acro{PF}{Particle Filter}
    \acro{LIB's}{Lithium-ion batteries}
    \acro{BMA}{Bayesian Model Averaging}
    \acro{NASA}{National Aeronautics and Space Administration}
    \acro{PCoE}{Prognostics Center of Excellence}
    \acro{FVS}{Feature Vector Selection}
    \acro{CS}{Cuckoo Search}
    \acro{LS-SVM}{Least Squares Support Vector Machine}
    \acro{SVM}{Support Vector Machine}
    \acro{SV}{Support Vector}
    \acro{SVR}{Support Vector Regression}
    \acro{ABC}{Artificial Bee Colony}
    \acro{ML}{Machine Learning}
    \acro{LSTM}{Long Short-Term Memory}
    \acro{RNN}{Recurrent Neural Network}
    \acro{DNN}{Deep Neural Network}
    \acro{ANN}{Artificial Neural Network}
    \acro{SOH}{State of Health}
    \acro{TCN}{Temporal Convolutional Network}
    \acro{AI}{Artificial Intelligence}
    \acro{SDE}{Stochastic Differential Equation}
    \acro{CPS}{Cyber-Physical Systems}
\end{acronym}

\vspace{0.15 cm}
\section*{Acknowledgment}
The research reported in this publication was supported by the Division of Research and Innovation at San Jos\'e State University under Award Number 23-UGA-08-044. The content is solely the responsibility of the author(s) and does not necessarily represent the official views of San Jos\'e State University.

\bibliographystyle{IEEEtran}
\bibliography{mybib}
\end{document}